\documentclass{article}
\PassOptionsToPackage{numbers,sort&compress}{natbib}

\usepackage[eandd, preprint]{neurips_2026}

\usepackage[utf8]{inputenc}
\usepackage[T1]{fontenc}
\usepackage{courier}
\usepackage{hyperref}
\usepackage{url}
\usepackage{booktabs}
\usepackage{tabularx}
\usepackage{array}
\usepackage{amsmath}
\usepackage{amssymb}
\usepackage{microtype}
\usepackage{xcolor}
\usepackage{graphicx}

\graphicspath{{../../docs/figures/generated/}}
\newcommand{\clip}{\operatorname{clip}}
\newcommand{\rankop}{\operatorname{rank}}
\newcommand{\indicator}[1]{\mathbf{1}\!\left[#1\right]}

\hypersetup{hidelinks}

\title{NeuroState-Bench: A Human-Calibrated Benchmark for Commitment Integrity in LLM Agent Profiles}

\author{
Xiao Jia\\
School of Artificial Intelligence\\
The Chinese University of Hong Kong, Shenzhen\\
\texttt{xiaojia@link.cuhk.edu.cn}
}

\date{}

\newcommand{\hcis}{\textsc{HCCIS-CORE}}
\newcommand{\hcisfull}{\textsc{HCCIS-FULL}}
\newcommand{\hcisn}{\textsc{HCCIS+N}}
\newcommand{\hnci}{\textsc{HNCI}}
\newcommand{\bgis}{\textsc{BGIS}}

\begin{document}

\maketitle

\begin{abstract}
Outcome-only evaluation under-specifies whether an evaluated agent profile preserves the commitments required to solve a multi-turn task coherently. NeuroState-Bench is a human-calibrated benchmark that operationalizes commitment integrity through benchmark-defined side-query probes rather than inferred hidden activations. The benchmark inventory contains 144 deterministic tasks and 306 benchmark-defined side-query probes spanning eight cognitively motivated failure families, paired clean and distractor variants, and three difficulty bands. The main 32-profile evaluation contains a fixed 16-profile local subset and a matched 16-profile hosted large-model subset evaluated through the same benchmark pipeline. Human calibration uses the final merged reporting scope: 104 sampled task units, 216 raw annotations, and 108 adjudicated task rows, with weighted $\kappa=0.977$ and $\mathrm{ICC}(2,1)=0.977$. Empirically, task success and commitment integrity diverge across this expanded grid: the success leader is not the integrity leader, 31 of 32 profiles change rank when integrity replaces task success, and integrity rankings are more stable under distractor perturbation. The primary confidence-free score HCCIS-CORE reaches $0.8469$ AUC and $0.6992$ PR-AUC for post-probe diagnostic discrimination of terminal task failure; the legacy full heuristic variant HCCIS-FULL reaches $0.7997$ AUC and $0.6410$ PR-AUC. Probe accuracy and state drift achieve slightly higher ROC-AUC ($0.8587$) and better Brier/ECE, while HCCIS-CORE has substantially higher point-estimate PR-AUC and remains more closely tied to the benchmark's intended construct. The exploratory neural-augmented variant HCCIS+N is weaker overall, and a randomized subspace control approaches chance. NeuroState-Bench therefore contributes a calibrated evaluation axis for exposing commitment failures over a broader model grid than the original local-only subset.
\end{abstract}

\section{Introduction}

Recent agent benchmarks have substantially improved the empirical study of tool use, web interaction, and long-horizon task completion, but they still overwhelmingly score agents by endpoint success \citep{workarena,osworld,taubench,longmemeval,membench,amabench}. That emphasis is useful when the only research question is whether a task was eventually solved. It is inadequate when the research question concerns how reliably an agent preserves the commitments that make successful behavior interpretable, robust, and auditable. In multi-turn settings, an agent may momentarily overwrite the wrong state, bind the wrong entity to the right value, preserve the right value under the wrong provenance, or fail to repair a contradicted belief, while still landing on the correct final answer by accident or by later recovery. Endpoint success alone does not reveal whether those cases reflect genuinely consistent operational commitment integrity.

NeuroState-Bench addresses that gap by treating operational commitment integrity as an evaluation target in its own right. Rather than attempting direct access to unobservable latent activations of deployed agents, it operationalizes commitment integrity as probeable commitment consistency between a task generator's gold commitment structure and full-prompt-conditioned commitment elicitation through benchmark-defined side-query probes. Probes are asked as separate side queries formed from the full task prompt plus a probe prompt; probe responses are not appended back into the task attempt used for final-answer scoring, and probe-location metadata functions as benchmark metadata and downstream weighting context rather than runtime transcript truncation. Under this operationalization, the benchmark measures whether elicited commitments remain consistent with the constraints, bindings, trusted sources, updates, and repaired beliefs associated with each probe. This definition gives an inspectable, reproducible measurement target without inferring mental states beyond the observable probe protocol.

NeuroState-Bench instantiates this target through eight task families, two variants, three difficulty bands, and deterministic instance generation, yielding 144 tasks and 306 side-query probes. A human calibration workflow contributes item-level difficulty, construct-validity status, and adjudicated annotation summaries. The evaluation covers a 32-profile grid that combines a fixed 16-profile local subset with a matched 16-profile hosted large-model subset over the same benchmark inventory. The evidence supports a benchmark-validity claim rather than a neural-prediction claim: task success systematically under-specifies whether an agent's full-prompt-conditioned elicited commitments remain consistent with task-defined commitments, and integrity-aware evaluation materially changes rankings, stability analyses, and failure interpretation. Empirically, the neural-augmented score does not dominate simpler baselines, so it is treated as an auxiliary comparator rather than the primary result.

Formally, let $\mathcal{T}$ denote the benchmark task set, let $P(t)$ denote the benchmark-defined side-query probes attached to task $t \in \mathcal{T}$, let $y_t \in \{0,1\}$ denote terminal task success, let $g_{t,p}$ denote the gold commitment associated with side-query probe $p \in P(t)$, and let $\hat g_{t,p}$ denote the normalized full-prompt-conditioned elicited probe response. The benchmark evaluates observable commitments rather than unobserved hidden states directly. It uses a probeable commitment-consistency relation
\begin{equation}
\mathcal{C}(t,p) = \indicator{\hat g_{t,p} = g_{t,p}},
\label{eq:commitment-consistency}
\end{equation}
where $\hat g_{t,p}$ is the agent's normalized full-prompt-conditioned elicited response to side-query probe $p$. Under this definition, outcome quality and operational commitment integrity are analytically distinct because
\begin{equation}
y_t \not\equiv \frac{1}{|P(t)|}\sum_{p \in P(t)} \mathcal{C}(t,p),
\label{eq:success-not-equivalent}
\end{equation}
and the empirical purpose of NeuroState-Bench is precisely to measure how often these quantities diverge.

The resulting contribution is an evaluation resource and accompanying analysis framework. Its core components are a cognitively structured task taxonomy, explicit side-query probe supervision for commitment assessment, a human-calibrated difficulty layer, ranking-validity analyses, phenotype discovery over agent families, and a delimited treatment of auxiliary neural and mechanistic signals. The \hnci{}, mechanistic-discovery, calibration, and \bgis{} modules function as supporting calibration context and comparators rather than as the central predictive claim.

\begin{figure*}[t]
\centering
\includegraphics[width=\textwidth]{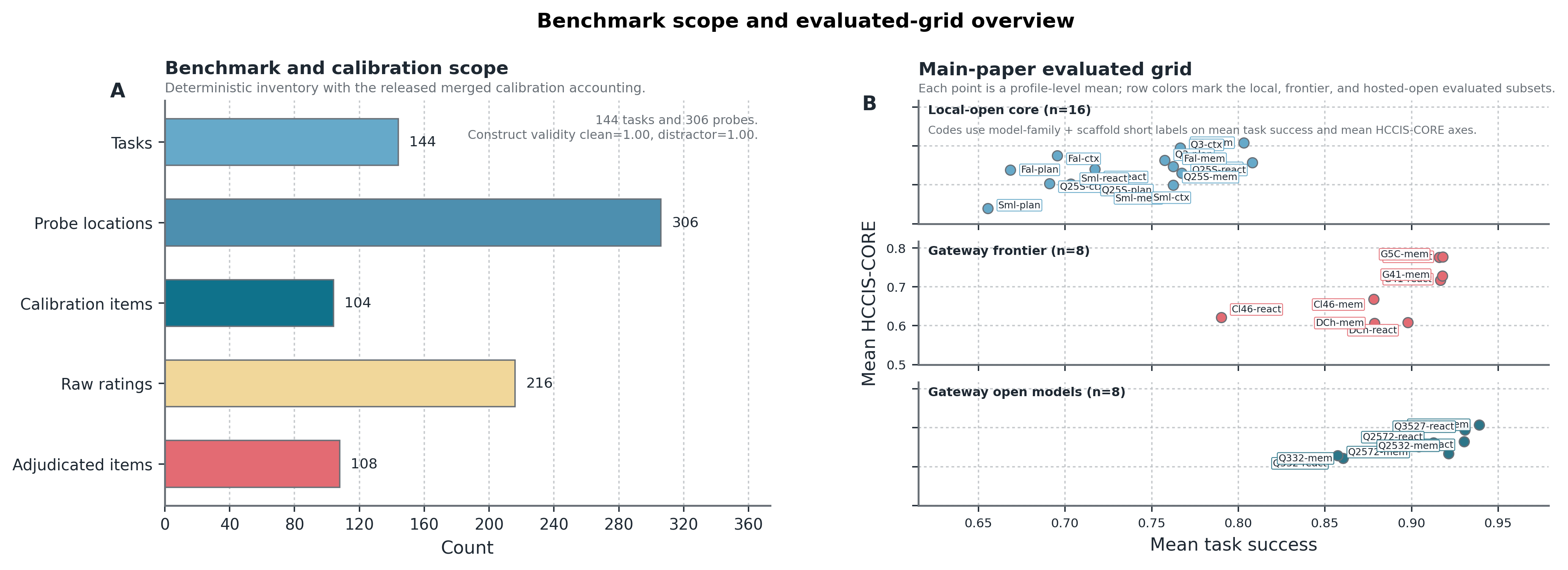}
\caption{Data-led overview of the 32-profile evaluated grid used in the primary analysis. Panel A summarizes deterministic benchmark scope and final merged calibration accounting, including 144 tasks, 306 benchmark-defined side-query probes, and the 104 / 216 / 108 sampled-raw-adjudicated counts. Panel B shows all 32 evaluated profiles directly as profile-level means on mean task-success and mean \hcis{} axes, using compact family-scaffold codes and the local-open, hosted-frontier, and hosted-open subset grouping; Appendix Table~\ref{tab:appendix-profile-codes-main} decodes those compact profile codes.}
\label{fig:overview}
\end{figure*}

\section{Related Work}

A first line of prior work evaluates agents through realistic task execution in external environments. WorkArena measures performance on knowledge-work tasks in a hosted enterprise platform \citep{workarena}; OSWorld evaluates multimodal agents in open computer environments \citep{osworld}; and $\tau$-bench studies tool-agent-user interaction in real-world domains \citep{taubench}. These benchmarks have made agent evaluation more realistic and more operationally relevant. However, their dominant outcome variable remains task completion. They are therefore well suited to studying end-to-end competence, but less suited to isolating whether an agent can maintain benchmark-defined commitments under explicit side-query elicitation.

A second line of work examines memory and long-horizon behavior more directly. LongMemEval, MemBench, and AMA-Bench all foreground memory as a limiting factor in agentic systems \citep{longmemeval,membench,amabench}. Those benchmarks are highly relevant to the present paper, because operational commitment integrity necessarily includes memory retention and state updating. NeuroState-Bench differs in two ways. First, it is explicitly probe-supervised: each task carries benchmark-defined side-query probes that query commitment consistency rather than infer it only from final task behavior. Second, it frames errors in terms of cognitively interpretable failure families such as constraint persistence, binding interference, source monitoring, contradiction repair, provenance binding, and stale-state overwrite.

A third line of work seeks cognitively or neuropsychologically grounded evaluation. NeuroCognition and CheeseBench exemplify an emerging interest in benchmarking large models against structured cognitive paradigms rather than unconstrained task collections \citep{neurocognition,cheesebench}. NeuroState-Bench draws on this direction but targets a different scientific object from general cognitive-ability benchmarks. Cognitive constructs are used instrumentally: they guide task-family design, distractor families, probe roles, and human calibration. This makes the benchmark interpretable without requiring strong anthropomorphic assumptions.

Neural and mechanistic components provide auxiliary analyses for constructing and validating conflict-sensitive evaluation signals. The \hnci{} pipeline extracts conflict-sensitive human signals from multiple datasets; the mechanistic pipeline searches for agreement-related components in small language models and calibrates them against human-side targets; and \bgis{} applies a neural-orientation mapping to probe-level features. These modules serve as calibration and validity context: they motivate calibration choices, document exploratory negative and mixed results, and define the evidence boundary for brain-to-agent analogies.

Because the present contribution is a benchmark rather than a new agent algorithm, it also draws on methodological traditions outside the immediate agent-benchmark literature. The design and calibration logic is explicitly shaped by construct validity and measurement theory \cite{cronbach1955construct}, inter-rater reliability analysis \cite{cohen1968weighted,shrout1979intraclass}, and calibration-aware predictive evaluation \cite{brier1950,fawcett2006roc,guo2017calibration}. These references ground the benchmark's measurement-theoretic claims. The central scientific question is whether the benchmark measures a coherent evaluation construct, not merely whether it yields another leaderboard.

Two additional literatures sharpen this positioning. First, newer agent benchmarks such as WebArena, AgentBench, GAIA, and SWE-bench push realism through browser environments, interactive tools, general-assistant tasks, and repository-level software engineering problems \cite{webarena,agentbench,gaia,swebench}. These resources motivate NeuroState-Bench but still primarily answer outcome questions such as whether a task was completed or an issue was resolved. NeuroState-Bench instead targets the commitment-consistency structure that those outcome metrics leave unmeasured. Second, process-supervision and reasoning-faithfulness work highlights the scientific value of richer process variables \cite{letsverify,faithfulcot,measuringcotfaithfulness}. Unlike process-supervision work, NeuroState-Bench does not supervise chain-of-thought text, but it shares the same evaluation-science motivation: final outcome alone is often too coarse to support causal or diagnostic interpretation.

This concern also connects to benchmark-validity discussions beyond agent evaluation. Recent work on benchmark contamination and benchmark trustworthiness emphasizes that an evaluation can fail scientifically even when its reported numbers are internally consistent \cite{benchmarkcontaminationsurvey}. One failure mode is contamination; another is construct mismatch. NeuroState-Bench addresses the latter by making operational commitment integrity the measured object directly rather than assuming it can be read off from endpoint success.

\section{NeuroState-Bench}

NeuroState-Bench is generated deterministically by code rather than assembled as a hand-curated spreadsheet. The main task builder enumerates eight families, two variants (clean and distractor), three difficulty bands (easy, medium, hard), and three instances per family--variant--band cell, yielding $8 \times 2 \times 3 \times 3 = 144$ tasks. Each task is assigned a family-specific cognitive construct, a deterministic seed, a gold state version, prompt metadata, and difficulty priors. The benchmark therefore exposes not only final prompts, but also the structured design variables from which they were produced. Table~\ref{tab:taxonomy} summarizes the task families.

At the inventory level, the benchmark structure can be written compactly as
\begin{equation}
|\mathcal{T}| = |\mathcal{F}|\,|\mathcal{V}|\,|\mathcal{D}|\,n_{\mathrm{inst}}
= 8 \cdot 2 \cdot 3 \cdot 3 = 144,
\label{eq:task-count}
\end{equation}
with $|\mathcal{F}|=8$ cognitive families, $|\mathcal{V}|=2$ prompt variants, $|\mathcal{D}|=3$ difficulty bands, and $n_{\mathrm{inst}}=3$ deterministic instances per cell. Because constraint persistence emits three probes for each of its 18 tasks while the remaining 126 tasks emit two probes each, the total probe inventory is
\begin{equation}
|\mathcal{P}| = 18 \cdot 3 + 126 \cdot 2 = 306.
\label{eq:probe-count}
\end{equation}
These counts are not editorial summaries; they are consequences of the generator's explicit combinatorial design and family-specific probe cardinalities.

\begin{table*}[t]
\centering
\scriptsize
\setlength{\tabcolsep}{3pt}
\caption{Released NeuroState-Bench task families. Probe counts are determined by the task generator; constraint persistence emits three probes per task, while the remaining families emit two.}
\label{tab:taxonomy}
\begin{tabularx}{\textwidth}{@{}>{\raggedright\arraybackslash}p{0.23\textwidth} >{\raggedright\arraybackslash}p{0.21\textwidth} >{\raggedright\arraybackslash}X >{\centering\arraybackslash}p{0.07\textwidth}@{}}
\toprule
Family & Cognitive construct & Core state requirement & \shortstack[c]{Probes\\per task} \\
\midrule
Constraint persistence & Goal maintenance & Preserve an early eligibility constraint while later candidate records foreground tempting but invalid alternatives. & 3 \\
Entity binding & Binding interference & Maintain the correct entity-to-slot assignment under near-miss swaps and roster-style confusion. & 2 \\
Conflict resolution & Source monitoring & Retain the trusted source when salient but explicitly untrusted sources conflict with it. & 2 \\
Delayed goal maintenance & Working-memory maintenance & Keep a delayed goal active across filler turns and late tempting replacements. & 2 \\
Tool-output distractor selection & External-interference control & Select the correct current tool output rather than stale or irrelevant outputs. & 2 \\
Repair after contradiction & Belief revision & Replace a contradicted state with the corrected one rather than preserving the stale belief. & 2 \\
Source-confusion provenance binding & Provenance binding & Preserve both content and provenance so that the correct value remains attached to the correct source. & 2 \\
Stale-state overwrite & State-replacement control & Overwrite obsolete state with an authoritative update despite repeated resurfacing of the stale state. & 2 \\
\bottomrule
\end{tabularx}
\end{table*}

\begin{table*}[t]
\centering
\footnotesize
\caption{Concrete benchmark item illustrating the task--probe schema. The example is a medium distractor constraint-persistence task that also appears in the automatically selected success-with-state-drift case studies.}
\label{tab:example-item}
\begin{tabularx}{\textwidth}{>{\raggedright\arraybackslash}p{0.24\textwidth} X}
\toprule
Component & Released example \\
\midrule
Family and construct & Constraint persistence; goal maintenance \\
Scenario & A dispatcher must preserve the original courier-eligibility gate while later notes foreground a near-valid courier. \\
Initial commitment & The active gate is ``verified badge + relay approval'' and the still-eligible courier is Rina. \\
Context turns & Rina satisfies both attributes; Remy lacks relay approval; Sora lacks badge verification; Ivo is on site but still in badge audit. \\
Distractor & A later override note says a temporary override should make courier Remy look eligible even though the original gate never changed. \\
Probe 1 & ``Under the original gate, which candidate is currently eligible?'' Gold: courier Rina. \\
Probe 2 & ``At the answer turn, what gate is still binding?'' Gold: verified badge + relay approval. \\
Probe 3 & ``If you answered now, which candidate should you still select?'' Gold: courier Rina. \\
Final matcher target & Final query: ``Which courier should still be selected under the original courier gate?'' Gold final answer: courier Rina. \\
Observed failure mode & For the SmolLM2 memory profile (compact code \texttt{Sml-mem}), the final answer is correct (``Courier Rina'') but probe accuracy is 0.0, state drift is 1.0, and \hcis{} is 0.1968. \\
\bottomrule
\end{tabularx}
\end{table*}

Each task prompt is programmatically composed from a fixed contract: family name, cognitive construct, difficulty band, answer contract, scenario, initial instruction, context turns, distractor description, and final user query. The benchmark is therefore not merely a set of prompts, but an explicit commitment-tracking environment in text form. Difficulty is an explicit design variable rather than an after-the-fact annotation. The generator seeds each difficulty band with human-difficulty and integrity-weight priors, delay length, contradiction density, source-conflict count, near-miss count, and tool-noise level. Those priors are subsequently merged with adjudicated human calibration where available. Unsampled items retain design priors, which is why the benchmark can remain fully materialized even when human calibration covers only a subset.

The task-construction code makes this difficulty layer explicit. Let $\omega_t^{\mathrm{design}} \in \{0.85, 1.00, 1.15\}$ denote the design-time integrity weight associated with the easy, medium, and hard bands respectively, and let $h_t \in [0,1]$ denote the human difficulty score after adjudicated overwrite where available. The merged item-level integrity weight used downstream is
\begin{equation}
\omega_t = \clip\!\left(\omega_t^{\mathrm{design}}\left(0.85 + 0.30 h_t\right),\,0.50,\,1.50\right).
\label{eq:item-weight}
\end{equation}
This definition is faithful to the calibration code and reflects an important design principle: human difficulty influences evaluation weight continuously rather than only through coarse band labels.

The benchmark's key methodological choice is to treat operational commitment integrity as probeable commitment consistency under benchmark-defined side-query probes. Every task carries an explicit probe specification list covering the probe prompt, location, construct, role, gold state version, and validity status. The benchmark contains 306 benchmark-defined side-query probes: the constraint-persistence family uses three probes per task to query early-constraint consistency more explicitly, whereas the other families use two. Probe scoring is family-aware. Responses are normalized and matched through the benchmark's answer matcher rather than through naive string equality, which is necessary because some tasks require value-only outputs, some require state labels, and provenance-binding tasks require structured source-value pairs.

The score family contains several closely related integrity scores. The primary analysis uses the confidence-free non-neural composition and refers to it as \hcis{}. It is not learned from held-out agent failures, and it is not a direct hidden-state readout. It is obtained by removing the confidence and uncertainty terms from the original non-neural heuristic and renormalizing the remaining non-confidence components, not by fitting weights to maximize terminal-failure AUC. Let $d_p$ denote probe drift, let $q_p = 1-c_p$ denote response-confidence risk, let $u_p$ denote the binary entropy of response confidence, let $t_p$ denote trajectory risk, let $x_p$ denote context-position risk, let $s_p$ denote success-probability risk, let $r_p$ denote retrieval risk, let $h_p$ denote the human-difficulty signal inherited through the merged task weights, and let $n_p$ denote the auxiliary neural-conflict proxy clipped to $[0,1]$. The entropy term is the normalized binary entropy \cite{shannon1948}
\begin{equation}
u_p = -c_p \log_2 c_p - (1-c_p)\log_2(1-c_p),
\label{eq:entropy-risk}
\end{equation}
and the context-position term is implemented as
\begin{equation}
x_p =
\clip\!\left(
0.45 z_p + 0.35 \phi_p + 0.20 \indicator{v_t=\mathrm{distractor}},
0,1
\right),
\label{eq:context-risk}
\end{equation}
where $z_p = \mathrm{pos}_p / \max_{q \in P(t)} \mathrm{pos}_q$ is the probe's normalized position within task $t$, $\phi_p \in \{0.15,0.50,0.85\}$ is the phase-risk schedule for initial, mid, and late probes, and $v_t$ is the task variant.

The primary probe-level failure risk is
\begin{equation}
\label{eq:probe-risk}
\rho^{\mathrm{core}}_p = \clip\!\left(0.7059d_p + 0.1471t_p + 0.1471x_p,\,0,\,1\right),
\end{equation}
and the probe-level integrity score is $\iota^{\mathrm{core}}_p = 1-\rho^{\mathrm{core}}_p$. The older full heuristic is retained as a legacy comparator, \hcisfull{},
\begin{equation}
\label{eq:probe-risk-full}
\rho^{\mathrm{full}}_p = \clip\!\left(0.48d_p + 0.18q_p + 0.14u_p + 0.10t_p + 0.10x_p,\,0,\,1\right),
\end{equation}
with integrity $\iota^{\mathrm{full}}_p = 1-\rho^{\mathrm{full}}_p$. The analysis tables retain separate primary-score and legacy-score columns for auditability. Human calibration enters the mainline score through the item weights in Equation~\ref{eq:item-weight}, the adjudicated validity weights, and the merged calibration tables rather than through an additional probe-level neural multiplier. Task-level integrity is then formed as a weighted aggregation of probe scores,
\begin{equation}
I_t = \frac{\sum_{p \in P(t)} w_p \iota_p}{\sum_{p \in P(t)} w_p},
\qquad
w_p = \clip\!\left(\alpha_p \omega_t \nu_p,\,0.10,\,3.00\right),
\label{eq:task-integrity}
\end{equation}
where $\iota_p$ denotes the integrity variant being aggregated, $\alpha_p$ is the probe weight, $\omega_t$ is the item-level integrity weight from Equation~\ref{eq:item-weight}, and $\nu_p$ is the probe-validity weight. Because $\omega_t$ is shared by all probes within task $t$, it cancels out of the task-level average whenever clipping is inactive; it becomes most important at the profile and dataset aggregation layers. For an agent profile $a$, aggregation is
\begin{equation}
I_a = \frac{\sum_{t \in \mathcal{T}} \Omega_t I_{a,t}}{\sum_{t \in \mathcal{T}} \Omega_t},
\qquad
\Omega_t = \omega_t,
\label{eq:profile-integrity}
\end{equation}
and dataset-level summaries use the same item-weighted aggregation over evaluated profile--task rows. In the evaluated configuration, the validity map uses weights 1.10 for adjudicated probes, 1.00 for probes validated during benchmark design, 0.95 for sampled probes, 0.90 for probes carrying only design-stage priors, and 0.75 for probes flagged as needing review. The auxiliary analysis also retains a neural-augmented exploratory variant, denoted here by \hcisn{}, with probe-level risk
\begin{equation}
\begin{aligned}
b_p &=
0.42d_p + 0.18q_p + 0.12u_p + 0.10t_p
+ 0.08x_p + 0.06s_p + 0.04r_p,\\
\tilde{\rho}_p
&= \clip\!\left(
b_p(0.80 + 0.35h_p + 0.15n_p),\,0,\,1
\right),
\end{aligned}
\label{eq:probe-risk-neural}
\end{equation}
and integrity $\tilde{\iota}_p = 1-\tilde{\rho}_p$. A randomized-subspace control is obtained by shuffling risk values. Across the benchmark configurations analyzed here, \bgis{} is disabled as a mainline evaluation path and retained only as a legacy exploratory comparator.

This score family is a diagnostic tool aligned with the benchmark's intended construct rather than a purely ex ante forecast. In the audit tables, state drift is a \emph{diagnostic post-probe} variable because it depends on the gold probe label after answer matching; response-confidence and uncertainty risks are available immediately after the elicited probe response but are excluded from the primary \hcis{} score; trajectory risk is available after the elicited probe response; and context-position risk depends only on benchmark metadata. By contrast, success-probability risk and retrieval risk are kept only in the exploratory comparator because they are more outcome-adjacent or optional diagnostic signals. This distinction constrains the diagnostic-discrimination results and avoids presenting \hcis{} as a general-purpose optimized predictor.

\begin{table}[t]
\centering
\footnotesize
\caption{Input and feature-availability summary for the primary benchmark outputs.}
\label{tab:side-query-semantics}
\begin{tabularx}{\columnwidth}{>{\raggedright\arraybackslash}p{0.26\columnwidth} >{\raggedright\arraybackslash}X >{\raggedright\arraybackslash}p{0.22\columnwidth}}
\toprule
Output or feature & Input used & Stage \\
\midrule
Final-answer scoring & Full task prompt and final response matched against the final gold answer & post-outcome \\
Probe-answer scoring & Full task prompt plus benchmark-defined side-query probe; probe response is not appended back into the task attempt & post-probe \\
\hcis{} & State drift, trajectory risk, and context-position risk & diagnostic post-probe \\
\hcisfull{} & \hcis{} terms plus response-confidence and uncertainty risks & legacy full heuristic \\
\hcisn{} & \hcisfull{}-style terms plus auxiliary neural, success-probability, and retrieval proxies & exploratory comparator \\
\bottomrule
\end{tabularx}
\end{table}

\section{Human Calibration and Auxiliary Neural Signals}

The benchmark's calibration layer is part of its scientific interpretation. The analysis uses the final merged calibration reporting scope rather than the earlier 96-item calibration pass. That merged scope contains 104 sampled task units, 216 raw annotations, and 108 adjudicated task rows. The merged sampled-item registry and the merged adjudicated scoring table are separate tables after final adjudication, so their row counts differ slightly. The merged-scope summary tables report weighted $\kappa=0.97696$, $\mathrm{ICC}(2,1)=0.97718$, a mean absolute inter-annotator difficulty gap of $0.0481$, and perfect agreement in the construct-label verification field. The benchmark also preserves scope-specific accounting for the initial calibration pass and the later focused recalibration pass, distinguishing accounting tables from final merged calibration tables rather than collapsing everything into a single denominator. This accounting makes the calibration denominator reproducible rather than opaque.

Let $r_{i,a} \in \{1,\dots,5\}$ denote annotator $a$'s difficulty rating for item $i$, and let $c_{i,a} \in \{0,1\}$ denote the construct-validity verification label for the same item. The normalized human difficulty score is produced by averaging ratings within item and dividing by the rating scale,
\begin{equation}
h_i = \clip\!\left(\frac{1}{5|A_i|}\sum_{a \in A_i} r_{i,a},\,0,\,1\right),
\label{eq:human-difficulty}
\end{equation}
where $A_i$ is the set of annotators available for item $i$. Pairwise consistency on a given reporting scope is then summarized by weighted Cohen's $\kappa$ \cite{cohen1968weighted} and $\mathrm{ICC}(2,1)$ \cite{shrout1979intraclass}. This is methodologically important because the benchmark uses human judgments not merely for descriptive reporting, but to calibrate difficulty and weighting fields. An independent blinded construct-assignment check covers 48 task--probe rows. Two annotators assigned each row to one of the eight benchmark families using only task and probe text, with task IDs, probe IDs, gold labels, and family/construct prompt headers hidden. Both annotators recovered the design family for all 48 rows, yielding per-annotator accuracy 1.000, macro accuracy 1.000, pairwise agreement 1.000, Cohen's $\kappa=1.000$, and no disagreement rows. This check establishes label-masked readability of the family distinctions without making an exhaustiveness claim about the construct set.

Human judgments behave in the direction required for a credible difficulty ladder. In the final merged calibration table, mean human difficulty is $0.4457$ for the easy band, $0.6086$ for the medium band, and $0.8118$ for the hard band, and the summary tables record the monotonicity check as passed. Distractor items are also harder for annotators than clean items in the calibration summaries. Construct-valid rates are 1.0 for both clean and distractor partitions in the sampled calibration rows, which means that the benchmark's family labels are not merely design aspirations but survive manual verification. Appendix Figure~\ref{fig:appendix-calibration-main} visualizes the calibration dashboard and makes the scope accounting explicit. Table~\ref{tab:summary} collects the most important inventory and calibration statistics.

\begin{table}[t]
\centering
\small
\caption{Benchmark inventory and calibration summary for the 32-profile evaluated grid and merged calibration reporting scope.}
\label{tab:summary}
\begin{tabularx}{\columnwidth}{>{\raggedright\arraybackslash}X >{\raggedleft\arraybackslash}p{0.34\columnwidth}}
\toprule
Statistic & Value \\
\midrule
Benchmark tasks & 144 \\
Probe locations & 306 \\
Families $\times$ variants $\times$ difficulty bands & $8 \times 2 \times 3$ \\
Instances per family--variant--band cell & 3 \\
Agent profiles in main 32-profile evaluated grid & 32 \\
Sampled calibration task units & 104 \\
Raw human annotations & 216 \\
Adjudicated merged-scope task rows & 108 \\
Weighted $\kappa$ for difficulty ratings & 0.977 \\
$\mathrm{ICC}(2,1)$ for difficulty ratings & 0.977 \\
Construct-verification Cohen's $\kappa$ & 1.000 \\
Blinded construct-assignment accuracy / $\kappa$ & 1.000 / 1.000 \\
Mean human difficulty (easy / medium / hard) & 0.446 / 0.609 / 0.812 \\
\bottomrule
\end{tabularx}
\end{table}

The auxiliary human-side pipeline is richer than the benchmark strictly requires. The \hnci{} workflow builds lexical, syntactic, surprisal, acoustic, and conflict regressors; runs ridge-style encoding models for fMRI; reuses temporal response modeling for ECoG; and produces event-, unit-, ROI-, and lag-level summaries. The analyzed outputs contain 218 event-level rows, 1{,}438 unit summaries, 14 ROI summaries, and 216 lag summaries across the Narratives, Podcast ECoG, and SMN4Lang resources. The control summaries show that these effects are present but small and heterogeneous: some fMRI ROI effects are on the order of $10^{-5}$ to $10^{-4}$, one ECoG ROI family is negative, and a targeted temporal window in the Podcast ECoG resource shows a larger positive effect. These outputs demonstrate that the human-side resource layer is empirically instantiated, structured, and analyzable, but do not support a strong universal neural predictor of agent failure.

At the implementation level, the encoding analyses are regularized predictive models of the form
\begin{equation}
\hat{\beta}_{\lambda}
=
\arg\min_{\beta}
\|y - X\beta\|_2^2 + \lambda \|\beta\|_2^2,
\label{eq:ridge}
\end{equation}
with downstream evaluation summarized in held-out $\Delta R^2$-style statistics \cite{hoerl1970ridge,naselaris2011encoding,crosse2016mtrf}. These results treat this pipeline as a quantitative auxiliary signal-generation layer with limited downstream utility, separate from the benchmark's main result.

The mechanistic and calibration layers show the same boundary. The mechanistic analysis searches for agreement-related component sets in GPT-2 and Pythia using predeclared campaigns, uses intervention-derived mechanistic features, and calibrates them to human-side targets using ridge regression with leave-one-group-out evaluation. Campaign selection is proxy-based rather than tuned on downstream benchmark outcomes; in the analyzed run, both selected campaigns are sparse separator-based component sets. The resulting held-out incremental validity is small or negative. The summary tables report a mechanistic mean $\Delta R^2$ of $-0.00197$, and the variance-partitioning tables show unique mechanistic contributions that are numerically tiny. In the notation of incremental predictive validity,
\begin{equation}
\Delta R^2 = R^2(X_{\mathrm{base}}, X_{\mathrm{mech}}) - R^2(X_{\mathrm{base}}),
\label{eq:delta-r2}
\end{equation}
and the results indicate that this increment is not reliably positive. The neural and mechanistic layers therefore serve as auxiliary calibration context rather than central evidence for the benchmark's value, as reflected in the analysis code and outputs.

\section{Experimental Protocol}

The implementation supports a broader serving interface than the original local benchmark core alone, including local execution and externally served large-model endpoints. The main 32-profile experiment therefore extends the older 16-profile local-only slice. This grid contains 32 profiles over the same 144-task benchmark inventory. The first half is a fixed local 16-profile subset obtained by crossing four base models---Qwen3-4B, Qwen2.5-1.5B, Falcon3-1B, and \mbox{SmolLM2-1.7B}---with react-style, context-engineered, planner-executor, and memory-retrieval wrappers. The second half is a matched hosted large-model subset obtained by crossing the react-style and memory-retrieval wrappers with GPT-4.1, GPT-5-Chat, Claude Sonnet 4.6, DeepSeek-Chat, Qwen2.5-32B-Instruct, Qwen2.5-72B-Instruct, Qwen3-32B, and Qwen3.5-27B. The expanded evaluation therefore no longer depends solely on the smaller local subset.

Both halves of the evaluated grid use the same deterministic task generator, the same probe specifications, the same answer matcher, and the same downstream scoring path. All runs in the main 32-profile experiment use deterministic decoding. The local subset is executed through the fixed local serving stack, while the larger-model half is executed through an OpenAI-compatible hosted path. The serving distinction affects reproducibility engineering and compute accounting, but not the measured construct, because the commitment-integrity computation is downstream of the shared task and probe outputs rather than specific to any single provider interface. The planner-executor wrapper continues to use the CLI adapter in the local subset, whereas the hosted profiles use OpenAI-compatible HTTP execution.

The fixed local 16-profile subset was executed on a single-machine setup with an NVIDIA GeForce RTX 5070 Laptop GPU reporting 8{,}151 MiB VRAM, an AMD Ryzen 9 8945HX CPU with 16 physical cores and 32 logical processors, and approximately 31.8 GiB system memory. The hardware summary contains four local server groups, all served with 4-bit loading. On this hardware, the 16-profile local subset required approximately one hour of wall-clock time. The additional 16 hosted profiles were executed through the same benchmark pipeline via a third-party OpenAI-compatible endpoint; provider-side accelerator configuration is not exposed by that endpoint. These details document reproducibility on commodity local hardware for the fixed local subset while supporting a broader 32-profile evaluation for external-validity testing.

The evaluation pipeline materializes task-level and probe-level outputs, then computes benchmark-derived and auxiliary summaries. At the task level, it reports task success, probe accuracy, state drift rate, repair precision, distractor susceptibility gap, the primary \hcis{} score, the exploratory neural-augmented variant \hcisn{}, and a collection of baseline or proxy terms. Probe accuracy and state drift are complements of one another in the benchmark implementation, so they provide equivalent ranking information but remain separately useful because one is a correctness view and the other is a risk view. Distractor susceptibility gap is computed as the difference between mean clean-condition integrity and mean distractor-condition integrity within a grouping. Post-probe diagnostic discrimination is evaluated against terminal task failure using AUC, PR-AUC, Brier score, and expected calibration error, with summaries reported overall, by family, by agent, and by agent-family slice. The code also supports bootstrap summaries and leave-agent-out reporting.

The reporting pipeline goes beyond ordinary bootstrap reporting. It computes task-block bootstrap, agent-block bootstrap, and two-way cluster bootstrap intervals so that uncertainty reflects the dependence induced by shared tasks and shared agents. It also reports paired bootstrap deltas comparing \hcis{} against \hcisn{}, probe accuracy, and state drift. Finally, it fits trained logistic and isotonic-calibrated baselines under leave-agent-out and leave-family-out protocols using the same benchmark-derived risk features. These analyses test the stability of the qualitative conclusions and make the auditability-versus-optimization trade-off explicit.

For a predictor producing task-level failure probabilities $\hat p_i$ against binary failure labels $y_i \in \{0,1\}$, the two calibration-sensitive summary statistics used throughout the analysis are
\begin{equation}
\mathrm{Brier} = \frac{1}{N}\sum_{i=1}^{N}(\hat p_i - y_i)^2,
\qquad
\mathrm{ECE} = \sum_{b=1}^{B}\frac{|B_b|}{N}\left|\mathrm{acc}(B_b)-\mathrm{conf}(B_b)\right|,
\label{eq:brier-ece}
\end{equation}
where $B=5$ bins in the default implementation, $\mathrm{acc}(B_b)$ is empirical failure frequency inside bin $b$, and $\mathrm{conf}(B_b)$ is mean predicted failure probability in that bin \cite{brier1950,guo2017calibration}. These metrics make calibration part of the evaluation, so risk scores are assessed not only by discrimination but also by their behavior as early-warning signals.

Ranking validity is treated as a first-class analysis target rather than as a side effect. The ranking module computes pairwise ranking disagreement across predictors, bootstrap rank stability, leave-family-out robustness, leave-agent-out robustness, and a misorder audit that compares the primary reference predictor (\hcis{}) to task success. The misorder audit is not a synthetic summary only; it can be joined back to task prompts, final answers, gold answers, and, where available, raw traces. This design supports item-level auditability: when integrity-aware rankings differ from success-based rankings, the audit records can be traced back to task prompts, final answers, gold answers, and raw traces where available.

The ranking analyses can be written compactly. Let $\psi$ denote a scoring rule and let $\rankop_{\psi}^{(v)}(a)$ denote the rank of agent $a$ under variant $v \in \{\mathrm{clean},\mathrm{distractor}\}$. The clean-versus-distractor instability score is
\begin{equation}
\mathrm{RI}_{\psi}
=
\frac{1}{A}
\sum_{a=1}^{A}
\left|
\rankop_{\psi}^{(\mathrm{clean})}(a) -
\rankop_{\psi}^{(\mathrm{distractor})}(a)
\right|,
\label{eq:ranking-instability}
\end{equation}
and the reference-versus-comparator mean absolute rank shift is
\begin{equation}
\mathrm{RS}_{\psi}
=
\frac{1}{A}
\sum_{a=1}^{A}
\left|
\rankop_{\hcis}(a) - \rankop_{\psi}(a)
\right|.
\label{eq:rank-shift}
\end{equation}
The ranking module also reports Spearman and Kendall rank correlations between these orderings \cite{spearman1904,kendall1938}, allowing the ranking-validity analysis to distinguish between isolated flips and globally reordered score geometry.

The phenotype analysis is equally explicit. The implementation first constructs a wide feature matrix by aggregating metrics over the family $\times$ difficulty-band $\times$ variant grid. In the expanded 32-profile experiment, the feature set includes task success, probe accuracy, state drift, \hcis{}, and human difficulty over all 32 evaluated profiles. The matrix is standardized, projected with two-dimensional PCA, and clustered with KMeans using the six-cluster configuration for the expanded grid. Stability is estimated by 25 bootstrap resamples and by leave-family-out coordinate shifts. Finally, case studies are selected automatically from benchmark outputs rather than hand-picked after inspection, with categories covering success-with-state-drift, baseline-missed warnings, and repair or distractor-overwrite failures.

Let $M$ denote the metric set, and define agent $a$'s feature vector as
\begin{equation}
x_a
=
\left[
\mu_{a,f,d,v}^{(m)}
\right]_{m \in M,\, f \in \mathcal{F},\, d \in \mathcal{D},\, v \in \mathcal{V}},
\label{eq:phenotype-vector}
\end{equation}
where $\mu_{a,f,d,v}^{(m)}$ is the mean of metric $m$ over all tasks for agent $a$ in family $f$, difficulty band $d$, and variant $v$. PCA is used only for the two-dimensional coordinates \cite{jolliffe2002pca}; KMeans is fit on the standardized full feature vectors by minimizing the within-cluster sum of squares \cite{macqueen1967},
\begin{equation}
\min_{\{\ell_a\},\{\mu_k\}}
\sum_{a=1}^{A}
\left\|
\widetilde{x}_a - \mu_{\ell_a}
\right\|_2^2,
\label{eq:kmeans}
\end{equation}
where $\widetilde{x}_a$ is the standardized feature vector and $\ell_a \in \{1,\dots,K\}$ is the assigned cluster label. Bootstrap stability is then summarized through pairwise coassignment rates
\begin{equation}
s_{ab}
=
\frac{1}{B}
\sum_{b=1}^{B}
\indicator{\ell_a^{(b)} = \ell_b^{(b)}},
\label{eq:coassign}
\end{equation}
and leave-family-out robustness through per-agent coordinate shifts
\begin{equation}
\Delta_{a,f}
=
\left|
z_{a,1} - z_{a,1}^{(-f)}
\right|,
\label{eq:lfo-shift}
\end{equation}
where $z_{a,1}$ is the first phenotype coordinate for agent $a$ and $z_{a,1}^{(-f)}$ is the same coordinate recomputed after holding out family $f$.

\section{Results}

The first result is that the benchmark's human calibration layer continues to behave like a valid difficulty scaffold after the evaluated grid is expanded. Difficulty rises monotonically from easy to medium to hard in the final merged calibration table, and the family-level performance tables show that agent behavior degrades most sharply in the families where commitment maintenance should be hardest. Constraint persistence is again the clearest example: across the 192 agent-task instances in each difficulty band of the 32-profile main experiment, task success falls from $0.4427$ in the easy band to $0.3906$ in the hard band, while mean \hcis{} falls from $0.3710$ to $0.3520$. Delayed goal maintenance shows a similar but milder pattern, with hard-band \hcis{} dropping to $0.5705$ from $0.6028$ in the easy band. By contrast, source-confusion provenance binding and stale-state overwrite remain comparatively strong across bands, with hard-band \hcis{} values around $0.82$--$0.83$. The pattern indicates that the benchmark is not uniformly saturated even once substantially stronger hosted models are added.

\begin{figure*}[t]
\centering
\includegraphics[width=\textwidth]{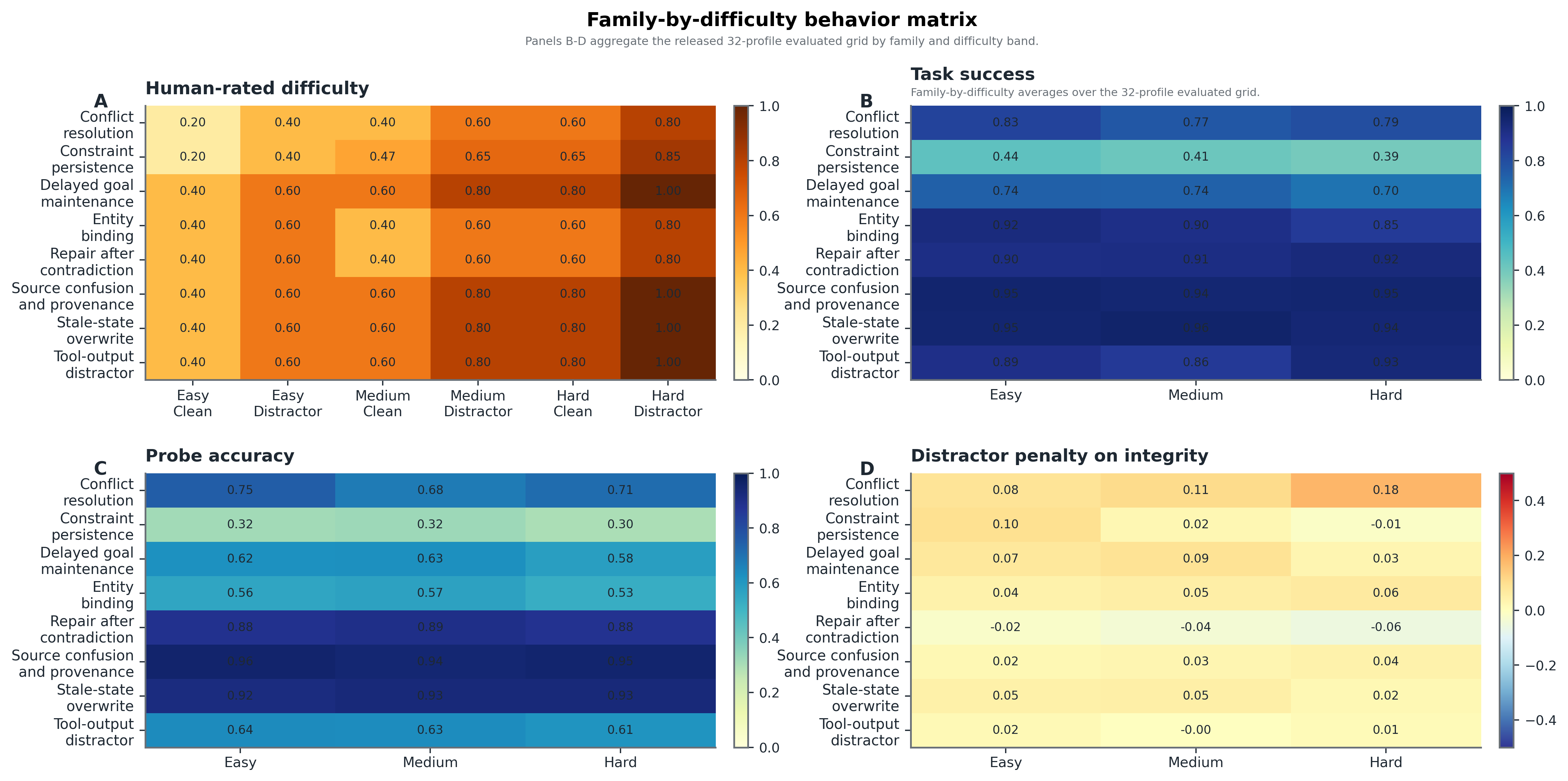}
\caption{Family-by-difficulty behavior matrix for the benchmark. Panel A visualizes the human-rated difficulty ladder over clean and distractor variants. Panels B--D report family-by-difficulty averages over the 32-profile evaluated grid: mean task success, mean probe accuracy, and the clean-minus-distractor integrity gap.}
\label{fig:task-matrix}
\end{figure*}

Figure~\ref{fig:task-matrix} makes this separation concrete at the family-by-difficulty level. The distractor penalty on integrity is not uniform across constructs. In this matrix, hard conflict resolution exhibits the largest positive distractor penalty ($0.20$), medium delayed-goal maintenance reaches $0.13$, and easy constraint persistence reaches $0.14$. By contrast, repair-after-contradiction shows near-zero or slightly negative distractor penalties (down to $-0.05$ in the medium band), indicating that its dominant failure mode in this evaluated grid is not simply distractor load but unsuccessful belief revision itself. Such construct-level variation would be lost in a single outcome-only aggregate.

The second result is that task success and commitment integrity remain strongly separable after the model grid is expanded. The task-success leader in the main 32-profile experiment is the Qwen3.5-27B memory profile (compact code \texttt{Q3527-mem}) at $0.9391$, but it is not the integrity leader. The highest \hcis{} score belongs to the GPT-5-Chat react profile (compact code \texttt{G5C-react}) at $0.7772$, followed by the GPT-5-Chat memory profile (\texttt{G5C-mem}) at $0.7758$ and the GPT-4.1 react profile (\texttt{G41-react}) at $0.7277$. In the legacy \hcisfull{} ordering, by contrast, the leading profiles remain the local Qwen3 variants. Relative to the primary \hcis{} axis, 31 of 32 evaluated profiles change rank when task success is used instead, with mean absolute rank shift $6.8125$, maximum shift $20$, Spearman correlation $0.5975$, and Kendall correlation $0.4153$. These mismatches are not artifacts of a single family. The agent-family breakdown shows that endpoint completion and elicited commitment consistency answer visibly different evaluative questions even when stronger hosted models improve both success and the primary core score.

A completed dual-annotator matcher audit over 193 stratified probe responses quantifies matcher error. The two annotators agreed on all 193 rows, all 97 automatic correct-match decisions were confirmed, but among 96 automatic incorrect-match decisions 45 were overturned as false negatives, meaning semantically correct answers rejected by the matcher, 36 were upheld as genuinely incorrect, and 15 were retained as unclear. Table~\ref{tab:appendix-matcher-audit-family-main} further shows that the unclear rows are concentrated in binding-heavy families, especially entity binding, whereas the false negatives are spread across several families rather than collapsing onto a single probe type. The verification artifacts also include a completed balanced follow-up audit pairing 128 hard-case automatic incorrect-match rows with 128 matched-correct controls matched as tightly as possible on audit group, family, variant, and response style. All 128 controls were confirmed, and most hard-case rows were overturned as false negatives or retained as unclear rather than upheld as genuinely incorrect; Tables~\ref{tab:appendix-matcher-audit-followup-main} and~\ref{tab:appendix-matcher-audit-followup-hardcases-main} report that follow-up audit separately from the primary overall audit. The matcher audit shows conservative under-matching, especially for verbose hosted responses; the sensitivity analysis therefore treats the reported integrity estimates as stress-tested but not matcher-free. Table~\ref{tab:appendix-matcher-sensitivity-main} bounds the primary-audit sensitivity directly: audit-row corrections keep the local-minus-hosted \hcis{} difference negative ($-0.0481$ to $-0.0445$) and leave 30--31 profiles rank-shifted against task success. Absolute top-5 membership is matcher-sensitive in some correction scenarios, while the broader success--integrity separation remains.

The phenotype outputs show the same separation at the behavior-space level. Appendix Figure~\ref{fig:appendix-phenotype-main} shows that the point with the highest success rate is not the point with the highest probe accuracy, and the largest integrity marker belongs to the Qwen3 memory variant rather than the Qwen3 react variant. Since the clustering is computed on the full standardized feature matrix rather than on the two-dimensional visualization alone, this discrepancy reflects the underlying benchmark structure rather than a projection artifact.

Post-probe diagnostic discrimination separates predictive utility from construct alignment. Table~\ref{tab:prediction} shows that the primary \hcis{} predictor reaches $0.8469$ AUC and $0.6992$ PR-AUC on terminal task failure in the expanded main experiment. Raw probe accuracy and state drift both reach slightly higher ROC-AUC ($0.8587$), better Brier/ECE, and lower PR-AUC ($0.5983$), while simpler proxy baselines remain near $0.40$ AUC and $0.15$ PR-AUC. The randomized subspace control rises only to $0.5193$ AUC, which is still close to chance in substantive terms and far below the structured score family. The legacy \hcisfull{} comparator reaches $0.7997$ AUC and $0.6410$ PR-AUC, and the exploratory neural-augmented comparator \hcisn{} reaches $0.7726$ AUC and $0.5693$ PR-AUC with worse Brier and ECE values than \hcis{}. In the expanded main experiment, the mainline score's value comes from structured aggregation and human calibration rather than cross-provider confidence terms or a demonstrated gain from the auxiliary neural term. The task-success predictor attains an AUC of 1.0 only because it is an oracle upper bound defined by the target label; it is therefore not an informative early-warning signal and is not comparable to diagnostic predictors derived directly from the benchmark in a substantive sense.

The cluster-aware uncertainty summaries quantify this pattern. Under task-block bootstrap, \hcis{} reaches $0.8481$ AUC with 95\% interval $[0.8111, 0.8799]$ and $0.7023$ PR-AUC with interval $[0.6182, 0.7724]$. Under the stricter two-way cluster bootstrap that simultaneously resamples tasks and agents, the same predictor reaches $0.8482$ AUC $[0.7815, 0.8923]$ and $0.7042$ PR-AUC $[0.5916, 0.7947]$. The corresponding two-way cluster intervals for \hcisn{} are $0.7746$ AUC $[0.6994, 0.8361]$ and $0.5747$ PR-AUC $[0.4484, 0.6821]$. These intervals qualify a simple point-estimate comparison: \hcis{} is a strong diagnostic score aligned with the benchmark's intended construct, while universal predictive dominance over probe accuracy or the neural term remains unsupported.

Paired clustered deltas against simpler comparators show a clearer pattern. Under two-way cluster bootstrap, the AUC delta between \hcis{} and probe accuracy is slightly negative, at $-0.0120$ with interval $[-0.0258, 0.0024]$, and the corresponding delta versus state drift is $-0.0120$ with interval $[-0.0303, 0.0022]$. Thus, the expanded experiment does not establish a raw ROC-AUC advantage for \hcis{}. By contrast, the PR-AUC deltas are positive at roughly $0.10$, and the comparison against the neural-augmented variant continues to favor \hcis{} cleanly on AUC, PR-AUC, Brier, and ECE. Accordingly, commitment-integrity scoring answers a different evaluative question rather than serving as a uniformly superior scalar predictor.

The predictor-gap outputs make the tradeoff explicit. Relative to \hcis{}, probe accuracy and state drift improve AUC slightly but lose about $0.10$ PR-AUC and produce different calibration behavior. The legacy \hcisfull{} variant is weaker than \hcis{} on all four overall diagnostic metrics, the exploratory neural-augmented variant \hcisn{} is weaker on AUC, PR-AUC, Brier score, and ECE, and the simpler uncertainty, success-proxy, trajectory, and retrieval baselines remain far worse overall. The mainline score's advantage is therefore interpretability, stronger precision-recall concentration than the direct correctness proxies, and ranking-validity evidence tied to the benchmark's intended construct, not dominance over every scalar discrimination metric.

\begin{table}[t]
\centering
\small
\caption{Diagnostic-discrimination performance for benchmark-derived predictors and controls. The task-success oracle is omitted from the main table because it is the complement of the terminal failure label by construction and therefore attains trivial perfect performance.}
\label{tab:prediction}
\begin{tabularx}{\columnwidth}{>{\raggedright\arraybackslash}X >{\raggedleft\arraybackslash}p{0.11\columnwidth} >{\raggedleft\arraybackslash}p{0.14\columnwidth} >{\raggedleft\arraybackslash}p{0.12\columnwidth} >{\raggedleft\arraybackslash}p{0.10\columnwidth}}
\toprule
Predictor & AUC & PR-AUC & Brier & ECE \\
\midrule
Human-calibrated commitment integrity (\hcis{}) & 0.8469 & 0.6992 & 0.1298 & 0.1657 \\
Legacy full heuristic (\hcisfull{}) & 0.7997 & 0.6410 & 0.1642 & 0.2223 \\
Exploratory neural-augmented variant (\hcisn{}) & 0.7726 & 0.5693 & 0.1864 & 0.2441 \\
Probe accuracy & 0.8587 & 0.5983 & 0.1297 & 0.1528 \\
State drift rate & 0.8587 & 0.5983 & 0.1297 & 0.1528 \\
Uncertainty proxy & 0.4009 & 0.1446 & 0.5869 & 0.5617 \\
Random-subspace control & 0.5193 & 0.1894 & 0.2377 & 0.2460 \\
\bottomrule
\end{tabularx}
\end{table}

\begin{figure*}[t]
\centering
\includegraphics[width=\textwidth]{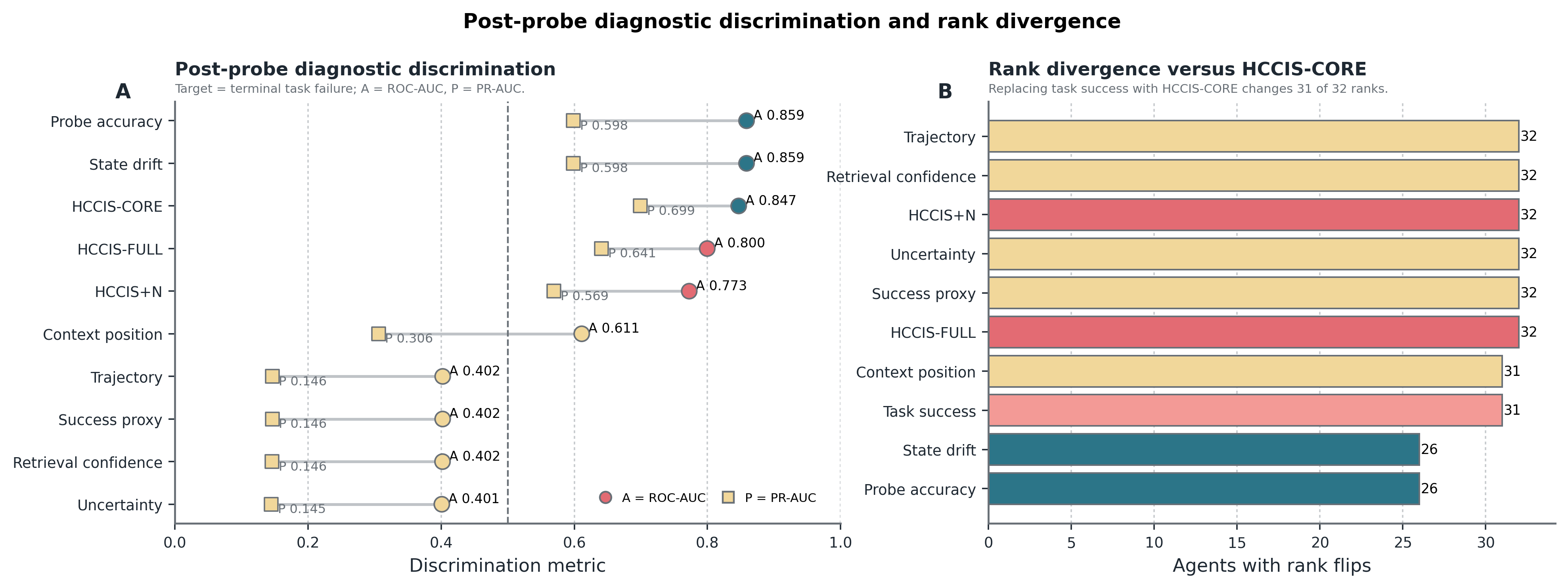}
\caption{Diagnostic discrimination and ranking divergence for the commitment-integrity axis. Panel A compares post-probe diagnostic discriminators of terminal task failure derived directly from the benchmark, with A denoting ROC-AUC and P denoting PR-AUC; in the expanded grid, probe accuracy and state drift achieve slightly higher ROC-AUC and better Brier/ECE than \hcis{}, while \hcis{} retains higher point-estimate PR-AUC and a closer connection to the benchmark's intended construct. Panel B reports rank disagreement against the commitment-integrity axis.}
\label{fig:failure}
\end{figure*}

Figure~\ref{fig:failure} also clarifies why the random-subspace, legacy-full, and neural-augmentation controls matter. These controls test whether the score exceeds arbitrary perturbations while remaining interpretable when auxiliary components are added or removed. The diagnostic tables show near-chance random-subspace performance and stronger confidence-free non-neural \hcis{} performance than the legacy full heuristic and neural-augmented comparator. Accordingly, neural calibration is treated as auxiliary rather than central.

The sensitivity analyses show that this conclusion is not brittle with respect to the primary \hcis{} weights, while also defining stability at the rank level. In the expanded grid, local $\pm 10\%$, local $\pm 25\%$, and Dirichlet perturbations do move some exact ranks relative to the primary ordering. Under local-$10\%$ perturbations, mean AUC is $0.8472$ with 5th and 95th percentiles $0.8465$ and $0.8483$, while mean absolute rank shift versus the primary ordering is only $0.3060$. Under local-$25\%$ perturbations, mean AUC is $0.8477$ with 5th and 95th percentiles $0.8443$ and $0.8523$, with mean absolute rank shift $0.8659$. Under Dirichlet resampling, mean AUC is $0.8485$ with 5th and 95th percentiles $0.8438$ and $0.8547$, with mean absolute rank shift $1.1187$. The corresponding mean numbers of profiles with any rank change are $8.4$, $15.0$, and $16.7$. These perturbation-induced shifts are still small compared with the larger task-success-versus-\hcis{} disagreement, whose mean absolute rank shift is $6.8125$.

The trained-baseline comparison adds a complementary perspective. Under leave-agent-out evaluation, the hand-specified \hcis{} reaches $0.8737$ AUC and $0.7625$ PR-AUC, exceeding the trained logistic and isotonic baselines on both discrimination metrics while remaining worse calibrated in Brier and ECE. Under leave-family-out evaluation, the picture changes: \hcis{} reaches $0.7648$ AUC and $0.5957$ PR-AUC, whereas the logistic audit-core baseline reaches $0.8381$ AUC and $0.5454$ PR-AUC. Thus, the hand-specified score is not universally predictive-optimal; its role is an auditable construct score that remains competitive on the precision-recall axis most relevant to a sparse failure target.

The feature-availability audit further separates diagnostic and ex ante uses of these scores. State drift is a post-probe diagnostic variable that uses the gold probe label after answer matching; response-confidence risk and uncertainty risk are available immediately after the probe response but are excluded from \hcis{}; context-position risk is determined entirely by benchmark metadata; and success-probability risk is deliberately excluded from the primary score because it is more outcome-adjacent. Terminal-task discrimination therefore shows that commitment integrity is related to downstream failure, not that the score is a purely ex ante early-warning detector. The confidence-comparability audit in Appendix Table~\ref{tab:appendix-confidence-main} extends this boundary: the old full heuristic and per-agent confidence-normalized variants induce substantial rank movement relative to the primary \hcis{} ordering. Confidence-related terms therefore function as legacy heuristic components rather than universally comparable uncertainty measurements across local and gateway profiles.

Appendix Figure~\ref{fig:appendix-controls-main} reports the negative-control diagnostics. In the variance-partitioning tables, shared variance exceeds $0.65$ in every reported row, while unique mechanistic contributions remain below $10^{-3}$ for GPT-2 and range from roughly $1.8 \times 10^{-4}$ to $6.9 \times 10^{-4}$ in four of the five reported Pythia control families. The one visibly larger Pythia baseline unique term in the primary-mechanism row reflects weaker mechanistic standalone fit rather than a positive mechanistic gain. Together, the controls support the evaluation-axis result while leaving neural augmentation outside the main score.

The ranking-validity analyses show why this difference matters. When agents are ranked by \hcis{} instead of task success, 31 of 32 evaluated profiles change rank, with mean absolute rank shift $6.8125$, maximum shift $20$, Spearman correlation $0.5975$, and Kendall correlation $0.4153$. These are not uniformly small perturbations around the same ordering. Family-specific rank flips are even more revealing. In the constraint-persistence family, the mean absolute shift between integrity and task-success rankings is $6.6875$, with 31 of the 32 agents changing position. In the conflict-resolution family, the mean absolute shift is $8.7188$, with 30 of the 32 agents flipping. These families place trusted commitments under central task-defining interference, making endpoint success least reliable as a sufficient summary of agent quality.

The benchmark also reveals that integrity-based ordering is more stable under the benchmark's clean-to-distractor perturbation than success-based ordering. The rank-instability summary reports a mean clean-versus-distractor rank shift of $4.3438$ for task success, compared with $2.7500$ for \hcis{}, with probe accuracy and state drift also at $2.7500$. Appendix Figure~\ref{fig:appendix-instability-main} provides the full family-level instability heatmap; repair-after-contradiction reaches the largest family-specific clean-versus-distractor shifts for probe accuracy and state drift, while conflict resolution and stale-state overwrite show large success-based instability despite relatively modest integrity instability.

The large-model half of the evaluated grid preserves the same qualitative separation outside the older small local subset, but the primary \hcis{} score changes the group-level interpretation relative to the legacy full heuristic. The hosted frontier subset attains slightly lower mean task success than the stronger hosted open-model subset ($0.8891$ versus $0.9070$) but higher mean \hcis{} ($0.6880$ versus $0.6577$). Both hosted subsets are more successful than the local subset and also higher on primary \hcis{} than the local subset mean of $0.6284$. By contrast, the legacy \hcisfull{} score retains the earlier local-higher pattern because it includes confidence and uncertainty terms whose cross-provider comparability is limited. The resulting geometry is not a fixed local-versus-hosted reversal: endpoint success, elicited commitment consistency, and confidence-sensitive heuristic variants induce different evaluation orderings.

The phenotype analysis adds another layer of structure. In the expanded evaluated grid, the PCA coordinates explain $23.25\%$ and $20.58\%$ of feature variance in the first two components, and the six-cluster configuration no longer collapses perfectly onto the original four local model families. The local families still occupy coherent regions, but the stronger hosted models partially co-cluster according to a shared high-success / moderate-integrity regime rather than by provider name alone. The resulting phenotype geometry suggests that broad behavioral structure in the benchmark is shaped jointly by base-model family and cross-family scale effects once stronger hosted models are added. NeuroState-Bench therefore does more than rank agents on a single axis; it exposes reproducible performance profiles over failure families and model regimes. Appendix Figures~\ref{fig:appendix-instability-main}, \ref{fig:appendix-phenotype-main}, \ref{fig:appendix-agent-breakdown-main}, and \ref{fig:appendix-agent-profiles-main} provide denser instability, phenotype, family-by-agent, and per-agent benchmark views.

Finally, the automatically selected case studies demonstrate that the benchmark captures hidden failures that would be missed by endpoint scoring alone even in the expanded grid. One state-drift case study uses the DeepSeek-Chat react profile (compact code \texttt{DCh-react}) on a hard distractor constraint-persistence item. The final answer is scored correct (``Bay 14''), but probe accuracy is $0.0$, state drift is $1.0$, and the primary \hcis{} score is only $0.1192$. A second case study uses the GPT-4.1 react profile (compact code \texttt{G41-react}) on a medium distractor constraint-persistence item; again the final answer is correct (``Packet D7''), yet probe accuracy is $0.0$, state drift is $1.0$, and \hcis{} is only $0.1192$. These examples are selected from saved benchmark outputs under a fixed case-study rule rather than manually chosen after inspection. They provide concrete benchmark instances in which endpoint success hides failed elicited commitments. Additional qualitative renderings are shown in Appendix Figure~\ref{fig:supp-cases}.

\section{Discussion}

NeuroState-Bench shows that commitment integrity is not interchangeable with endpoint success. It exposes cases where agents answer correctly despite state drift, reorders agent profiles, and yields more stable integrity rankings under distractor perturbation. The practical implication is that endpoint success should not be treated as a sufficient summary of agent reliability when commitments, provenance, constraints, or stale information matter. Reporting success together with \hcis{} and probe-level summaries makes failures inspectable at the task and family levels rather than collapsed into aggregate error rates.

NeuroState-Bench functions as a controlled diagnostic layer alongside broader capability evaluations. It compares wrappers, model families, and serving interfaces under the same conflict-sensitive task inventory, and it supports audits of whether ranking differences are associated with endpoint accuracy, probe consistency, state drift, or the confidence and uncertainty terms retained only in \hcisfull{}. The primary score is confidence-free because the results support that benchmark composition, while \hcisfull{}, \hcisn{}, and the neural and mechanistic layers serve as auxiliary comparators, construct checks, and controls.

The scope remains bounded. Probes are side queries over the full prompt, so they measure elicited commitment consistency rather than hidden-state preservation. The 32-profile grid is not a model census; tasks are synthetic and deterministic; and the core score is engineered. The benchmark exposes hidden failures as a diagnostic stress test, not as a deployment certificate or general safety guarantee. Public deterministic tasks may also invite benchmark gaming.

\section{Conclusion}

NeuroState-Bench is a human-calibrated benchmark for commitment integrity in LLM agent profiles. The benchmark's main contribution is methodological and evaluative rather than architectural: it provides a structured way to measure whether an agent's elicited commitments remain consistent with task-defined commitments, not merely whether it reached the correct endpoint. The results show that this distinction matters empirically. Task success and integrity diverge, rankings reorder, perturbation stability changes, and automatically extracted case studies reveal hidden failures that endpoint metrics miss.

The results also delimit the current evidence. Auxiliary neural and mechanistic modules are implemented and scientifically useful, but they do not justify making the neural-augmented variant the benchmark's primary score. NeuroState-Bench therefore offers a concrete and reproducible contribution to evaluation science for agentic AI.

The accompanying artifacts are also reproducibility-oriented. They provide deterministic task generation, probe specifications, adjudicated calibration tables, task- and probe-level benchmark outputs, phenotype coordinates, rank-flip audits, and supplementary figure inventories. As a result, the benchmark can be inspected not only at the level of aggregate performance, but at the level of individual probes, tasks, and ranking reversals. That granularity is essential if operational commitment integrity is to become a scientifically useful evaluation target rather than an informal construct.

\appendix

\section{Appendix: Supplementary Artifact Details}

The main 32-profile experiment materializes 4{,}608 agent-task evaluations (32 agents over 144 tasks) and retains explicit benchmark metadata for benchmark generation, human calibration, ranking validity, phenotype analysis, and case-study extraction. The agent grid combines a fixed 16-profile local subset with a matched 16-profile hosted large-model subset executed through the same benchmark inventory and scoring path. The local half is factored into four base models and four wrappers, which is important for interpretation because wrapper-level differences should not be mistaken for differences between independently trained agent systems. The hosted half extends the grid with two wrappers over eight additional large-model profiles. The resulting evaluation covers 4{,}608 agent-task rows and provides the empirical basis for the appendix analyses.

Table~\ref{tab:appendix-gateway-main} summarizes this final expanded evaluated grid directly. The distinction that still matters is not whether the large-model half is ``supplementary,'' but that the local subset is fully local whereas the additional 16 large-model profiles are served through a third-party OpenAI-compatible endpoint. That distinction affects compute reporting and artifact organization, but not the benchmark inventory against which all 32 profiles are compared.

\begin{table}[t]
\centering
\footnotesize
\setlength{\tabcolsep}{3pt}
\caption{Expanded evaluation summary used in the primary analysis. The 32-profile grid combines the fixed 16-profile local subset with 16 additional hosted large-model profiles over the same 144-task benchmark.}
\label{tab:appendix-gateway-main}
\begin{tabularx}{\columnwidth}{>{\raggedright\arraybackslash}X >{\raggedleft\arraybackslash}p{0.28\columnwidth}}
\toprule
Quantity & Value \\
\midrule
Evaluated profiles & 32 \\
Agent-task evaluations & 4{,}608 \\
Local subset + hosted subset & 16 + 16 \\
Success leader & \shortstack[r]{Qwen3.5-27B memory\\profile (0.9391)} \\
Integrity leader & \shortstack[r]{GPT-5-Chat react\\profile (0.7772)} \\
Profiles changing rank under \hcis{} vs task success & 31 / 32 \\
Mean absolute rank shift & 6.8125 \\
Maximum rank shift & 20 \\
Mean task success, hosted frontier subset & 0.8891 \\
Mean \hcis{}, hosted frontier subset & 0.6880 \\
Mean task success, stronger hosted open-model subset & 0.9070 \\
Mean \hcis{}, stronger hosted open-model subset & 0.6577 \\
Mean task success, local subset & 0.7338 \\
Mean \hcis{}, local subset & 0.6284 \\
\bottomrule
\end{tabularx}
\end{table}

Table~\ref{tab:appendix-profile-codes-main} decodes the compact family-scaffold codes used in Figure~\ref{fig:overview} and the appendix per-agent views so readers can move directly between compact figure labels and profile descriptions.

\begin{table*}[t]
\centering
\scriptsize
\setlength{\tabcolsep}{4pt}
\caption{Compact-code lookup for the 32 evaluated profiles shown in Figure~\ref{fig:overview} and Appendix Figures~\ref{fig:appendix-agent-breakdown-main} and~\ref{fig:appendix-agent-profiles-main}.}
\label{tab:appendix-profile-codes-main}
\begin{tabularx}{\textwidth}{>{\raggedright\arraybackslash}p{0.12\textwidth} >{\raggedright\arraybackslash}X >{\raggedright\arraybackslash}p{0.12\textwidth} >{\raggedright\arraybackslash}X}
\toprule
Code & Profile description & Code & Profile description \\
\midrule
\texttt{Q3-ctx} & Qwen3-4B context profile & \texttt{G41-mem} & GPT-4.1 memory profile \\
\texttt{Q3-mem} & Qwen3-4B memory profile & \texttt{G41-react} & GPT-4.1 react profile \\
\texttt{Q3-plan} & Qwen3-4B planner profile & \texttt{G5C-mem} & GPT-5-Chat memory profile \\
\texttt{Q3-react} & Qwen3-4B react profile & \texttt{G5C-react} & GPT-5-Chat react profile \\
\texttt{Q25S-ctx} & Qwen2.5-1.5B context profile & \texttt{Cl46-mem} & Claude Sonnet 4.6 memory profile \\
\texttt{Q25S-mem} & Qwen2.5-1.5B memory profile & \texttt{Cl46-react} & Claude Sonnet 4.6 react profile \\
\texttt{Q25S-plan} & Qwen2.5-1.5B planner profile & \texttt{DCh-mem} & DeepSeek-Chat memory profile \\
\texttt{Q25S-react} & Qwen2.5-1.5B react profile & \texttt{DCh-react} & DeepSeek-Chat react profile \\
\texttt{Sml-ctx} & SmolLM2 context profile & \texttt{Q2532-mem} & Qwen2.5-32B memory profile \\
\texttt{Sml-mem} & SmolLM2 memory profile & \texttt{Q2532-react} & Qwen2.5-32B react profile \\
\texttt{Sml-plan} & SmolLM2 planner profile & \texttt{Q2572-mem} & Qwen2.5-72B memory profile \\
\texttt{Sml-react} & SmolLM2 react profile & \texttt{Q2572-react} & Qwen2.5-72B react profile \\
\texttt{Fal-ctx} & Falcon3 context profile & \texttt{Q332-mem} & Qwen3-32B memory profile \\
\texttt{Fal-mem} & Falcon3 memory profile & \texttt{Q332-react} & Qwen3-32B react profile \\
\texttt{Fal-plan} & Falcon3 planner profile & \texttt{Q3527-mem} & Qwen3.5-27B memory profile \\
\texttt{Fal-react} & Falcon3 react profile & \texttt{Q3527-react} & Qwen3.5-27B react profile \\
\bottomrule
\end{tabularx}
\end{table*}

Auxiliary outputs further document the calibration context. On the human side, the \hnci{} outputs contain 218 event scores, 1{,}438 unit summaries, 14 ROI summaries, and 216 lag summaries, while the summaries record 12 SMN4Lang discovery subjects. On the mechanistic side, campaign search reports, selection tables, intervention outputs, feature tables, held-out calibration metrics, fitted mappings, and variance-partitioning tables are reported. These outputs make the auxiliary layers analyzable even where downstream predictive gains are small or absent.

The public artifacts cover the canonical benchmark resources: benchmark tables for the 32-profile evaluation, adjudicated calibration outputs, schema documentation, a benchmark card, a validity statement, and figure and table files, while excluding raw traces and upstream human/neural source data. The verification artifacts are broader. They include annotation instructions, the rating-rubric protocol, the anonymized raw dual-annotator rating matrix used to recompute weighted $\kappa$ and $\mathrm{ICC}(2,1)$, the completed blinded construct-assignment materials, raw task/probe traces, the ranking-trace audit, the completed primary stratified matcher-audit materials, the strict-schema side-query rerun, the balanced follow-up matcher audit, and hard-case and matched-correct control materials kept distinct from the primary overall audit. This split supports reproducibility while keeping the public benchmark artifacts scoped to the canonical benchmark resources.

Two artifact conventions are explicit. First, the earlier 96-item calibration pass is preserved separately from the final merged reporting scope used by the primary analysis and Figure~\ref{fig:overview}; exact table inventories are listed in the artifact inventories. Second, aggregate summary tables use an explicit pooled-row sentinel; this convention does not denote a 33rd evaluated profile.

The primary dual-annotator matcher audit covers 193 stratified raw probe responses together with the two completed annotator files, the merged adjudication table, and a summary table. Row-level agreement in this audit set is 1.0. All 97 automatic correct-match decisions were confirmed, whereas the 96 automatic incorrect-match decisions split into 36 human-confirmed incorrect cases, 45 false negatives, meaning semantically correct answers rejected by the matcher, and 15 unclear cases. Table~\ref{tab:appendix-matcher-audit-main} further shows that clear-row matcher accuracy is lower on verbose outputs ($0.6489$) than on concise outputs ($0.8571$), and somewhat lower on hosted rows ($0.7295$) than on the local subset ($0.7857$). Table~\ref{tab:appendix-matcher-audit-family-main} adds the family-level split for those 96 automatic incorrect-match rows: the 15 unclear cases are concentrated in entity binding and source/provenance binding, whereas the overturned false negatives are distributed across constraint persistence, repair-after-contradiction, and source/provenance binding. A separate strict-schema side-query rerun over the same 193 primary-audit rows completed for all rows under the original-agent policy. It required a structured answer response and matched only the parsed answer. Table~\ref{tab:appendix-strict-schema-main} reports a parse-valid rate of 0.9223 and a strict-schema match rate of 0.5078 overall; hosted rows parse cleanly in all cases, while local rows have lower JSON compliance. Because this audit set is stratified and enriched for automatic correct-match and incorrect-match audit rows, the strict-schema match rate estimates schema robustness rather than full-benchmark strict-schema accuracy. The rerun provides schema-feasibility evidence alongside the raw-output primary benchmark. A separate balanced follow-up audit pairs the 128 hard-case automatic incorrect-match rows from the second-round review set with 128 matched-correct controls sampled without overlap from the remaining probe inventory. All 128 controls were confirmed, whereas the hard-case rows split into 87 false negatives, 23 upheld incorrect cases, and 18 unclear cases. Table~\ref{tab:appendix-matcher-audit-followup-main} summarizes the completed follow-up audit, and Table~\ref{tab:appendix-matcher-audit-followup-hardcases-main} breaks down the hard-case adjudications by family and response style. Control matching is exact on audit group, family, variant, and response style for 104 of the 128 controls, relaxes one level for 2 rows, and relaxes to family plus response style for the remaining 22. The completed primary audit provides independent validation of the outcome-versus-integrity separation, while the strict-schema rerun and completed follow-up audit provide robustness checks without treating the audits as a pooled overall-accuracy estimate.

\begin{table}[t]
\centering
\footnotesize
\caption{Completed dual-annotator matcher audit for the expanded 32-profile experiment. Here ``false negative'' denotes an automatic incorrect-match decision overturned by manual review, and ``clear-row matcher accuracy'' excludes rows adjudicated as unclear.}
\label{tab:appendix-matcher-audit-main}
\begin{tabularx}{\columnwidth}{>{\raggedright\arraybackslash}X >{\raggedleft\arraybackslash}p{0.28\columnwidth}}
\toprule
Matcher-audit quantity & Value \\
\midrule
Sampled probe responses & 193 \\
Annotator agreement & 193 / 193 (1.000) \\
Automatic correct-match decisions confirmed & 97 / 97 (1.000) \\
Automatic incorrect-match decisions overturned as false negatives & 45 / 96 (0.469) \\
Automatic incorrect-match decisions upheld as genuinely incorrect & 36 / 96 (0.375) \\
Automatic incorrect-match decisions retained as unclear & 15 / 96 (0.156) \\
Clear-row matcher accuracy, local subset & 0.7857 \\
Clear-row matcher accuracy, gateway-combined rows & 0.7295 \\
Clear-row matcher accuracy, concise responses & 0.8571 \\
Clear-row matcher accuracy, verbose responses & 0.6489 \\
\bottomrule
\end{tabularx}
\end{table}

\begin{table}[t]
\centering
\footnotesize
\caption{Family-level adjudication breakdown for the 96 automatic incorrect-match decisions in the completed matcher audit. Counts sum to the automatic incorrect-match total reported in Table~\ref{tab:appendix-matcher-audit-main}.}
\label{tab:appendix-matcher-audit-family-main}
\begin{tabularx}{\columnwidth}{>{\raggedright\arraybackslash}X >{\raggedleft\arraybackslash}p{0.12\columnwidth} >{\raggedleft\arraybackslash}p{0.16\columnwidth} >{\raggedleft\arraybackslash}p{0.12\columnwidth}}
\toprule
Family & False neg. & Upheld incorrect & Unclear \\
\midrule
Conflict resolution & 4 & 6 & 0 \\
Constraint persistence & 12 & 2 & 2 \\
Delayed goal maintenance & 7 & 1 & 0 \\
Entity binding & 4 & 2 & 10 \\
Repair after contradiction & 8 & 7 & 0 \\
Source/provenance binding & 5 & 1 & 3 \\
Stale-state overwrite & 3 & 11 & 0 \\
Tool-output distractor selection & 2 & 6 & 0 \\
\bottomrule
\end{tabularx}
\end{table}

\begin{table*}[t]
\centering
\footnotesize
\caption{Strict-schema side-query rerun over the completed 193-row primary matcher-audit subset. The subset is stratified and enriched for automatic matcher-audit decisions, so the match rate estimates schema robustness rather than full-benchmark strict-schema accuracy. The rerun uses original agent endpoints/models, requires a structured answer response, and matches only the parsed answer.}
\label{tab:appendix-strict-schema-main}
\begin{tabularx}{\textwidth}{>{\raggedright\arraybackslash}X >{\raggedleft\arraybackslash}p{0.10\textwidth} >{\raggedleft\arraybackslash}p{0.12\textwidth} >{\raggedleft\arraybackslash}p{0.16\textwidth} >{\raggedleft\arraybackslash}p{0.18\textwidth}}
\toprule
Stratum & Rows & Completed & Parse-valid rate & Strict-schema match rate \\
\midrule
Overall & 193 & 193 & 0.9223 & 0.5078 \\
Gateway frontier & 79 & 79 & 1.0000 & 0.5316 \\
Gateway open extension & 54 & 54 & 1.0000 & 0.5370 \\
Local open core & 60 & 60 & 0.7500 & 0.4500 \\
Concise responses & 92 & 92 & 0.9239 & 0.5326 \\
Verbose responses & 101 & 101 & 0.9208 & 0.4851 \\
Original correct-match rows & 97 & 97 & 0.9381 & 0.6289 \\
Original incorrect-match rows & 96 & 96 & 0.9063 & 0.3854 \\
\bottomrule
\end{tabularx}
\end{table*}

\begin{table*}[t]
\centering
\footnotesize
\caption{Matcher sensitivity bound over the completed 193-row primary audit. Corrections are applied only to audited probe-answer rows and are not pooled with the hard-case follow-up audit. Local/hosted values report mean \hcis{}; rank shifts compare \hcis{} against task-success rankings over the 32-profile grid.}
\label{tab:appendix-matcher-sensitivity-main}
\begin{tabularx}{\textwidth}{>{\raggedright\arraybackslash}X >{\raggedleft\arraybackslash}p{0.13\textwidth} >{\raggedleft\arraybackslash}p{0.15\textwidth} >{\raggedleft\arraybackslash}p{0.08\textwidth} >{\raggedleft\arraybackslash}p{0.12\textwidth} >{\raggedleft\arraybackslash}p{0.12\textwidth} >{\raggedleft\arraybackslash}p{0.08\textwidth}}
\toprule
Scenario & Corrected audit rows & Local / hosted \hcis{} & Gap & Rank shift mean / max & $\rho$ / $\tau$ & Top-5 changed \\
\midrule
Automatic matcher & 0 & 0.6284 / 0.6728 & $-0.0445$ & 6.8125 / 20 & 0.5975 / 0.4153 & Ref. \\
Clear false negatives corrected & 45 & 0.6302 / 0.6771 & $-0.0469$ & 6.5625 / 20 & 0.6250 / 0.4395 & Yes \\
Unclear $\rightarrow$ correct & 60 & 0.6308 / 0.6789 & $-0.0481$ & 6.5000 / 20 & 0.6312 / 0.4435 & No \\
Unclear $\rightarrow$ incorrect & 45 & 0.6302 / 0.6771 & $-0.0469$ & 6.5625 / 20 & 0.6250 / 0.4395 & Yes \\
\bottomrule
\end{tabularx}
\end{table*}

\begin{table}[t]
\centering
\footnotesize
\caption{Completed balanced follow-up matcher audit kept separate from the primary 193-row overall audit. The completed audit pairs the 128 hard-case automatic incorrect-match rows from the second-round review set with 128 matched-correct controls sampled without overlap from the remaining probe inventory.}
\label{tab:appendix-matcher-audit-followup-main}
\begin{tabularx}{\columnwidth}{>{\raggedright\arraybackslash}X >{\raggedleft\arraybackslash}p{0.28\columnwidth}}
\toprule
Follow-up quantity & Value \\
\midrule
Completed balanced follow-up rows & 256 \\
Hard-case automatic incorrect-match rows & 128 \\
Matched-correct controls confirmed & 128 / 128 (1.000) \\
Hard-case rows overturned as false negatives & 87 / 128 (0.680) \\
Hard-case rows upheld as genuinely incorrect & 23 / 128 (0.180) \\
Hard-case rows retained as unclear & 18 / 128 (0.141) \\
Control overlap with primary 193-row audit & 0 \\
Control overlap with hard-case negative batch & 0 \\
Exact matches on \shortstack[r]{audit group + family +\\variant + response style} & 104 / 128 \\
Fallback matches on \shortstack[r]{audit group + family +\\response style} & 2 / 128 \\
Fallback matches on family + response style & 22 / 128 \\
\bottomrule
\end{tabularx}
\end{table}

\begin{table}[t]
\centering
\footnotesize
\caption{Completed adjudication breakdown for the 128 hard-case automatic incorrect-match rows in the balanced follow-up audit. The response-style rows aggregate the same 128 hard-case rows rather than adding new cases.}
\label{tab:appendix-matcher-audit-followup-hardcases-main}
\begin{tabularx}{\columnwidth}{>{\raggedright\arraybackslash}X >{\raggedleft\arraybackslash}p{0.12\columnwidth} >{\raggedleft\arraybackslash}p{0.12\columnwidth} >{\raggedleft\arraybackslash}p{0.12\columnwidth}}
\toprule
Slice & False neg. & Upheld incorrect & Unclear \\
\midrule
Constraint persistence & 43 & 12 & 1 \\
Entity binding & 32 & 0 & 8 \\
Source/provenance binding & 12 & 11 & 9 \\
\midrule
Concise hard-case rows & 16 & 16 & 11 \\
Verbose hard-case rows & 71 & 7 & 7 \\
\bottomrule
\end{tabularx}
\end{table}

\begin{table}[t]
\centering
\footnotesize
\caption{Compute resources for the expanded 32-profile experiment. The local rows describe the fixed 16-profile local subset; the gateway row clarifies the serving boundary for the additional 16 large-model profiles.}
\label{tab:appendix-compute-main}
\begin{tabularx}{\columnwidth}{>{\raggedright\arraybackslash}X >{\raggedleft\arraybackslash}p{0.42\columnwidth}}
\toprule
Resource & Value \\
\midrule
Local GPU & NVIDIA GeForce RTX 5070 Laptop GPU \\
Reported VRAM & 8{,}151 MiB \\
CPU & AMD Ryzen 9 8945HX \\
CPU topology & 16 physical cores / 32 logical processors \\
System memory & $\approx 31.8$ GiB \\
Local server groups & 4 \\
Local model loading & 4-bit loading for all four local server groups \\
Decoding & deterministic decoding across the full evaluated grid \\
Approximate local runtime & $\approx 1.0$ hour for the 16-profile local subset \\
Hosted large-model half & 16 externally served profiles via one OpenAI-compatible endpoint; provider-side accelerator details not exposed \\
\bottomrule
\end{tabularx}
\end{table}

\begin{table*}[t]
\centering
\scriptsize
\caption{Cluster-aware uncertainty summaries for the main diagnostic-discrimination comparisons. Task-block bootstrap resamples tasks while preserving all agent evaluations for each task; agent-block bootstrap resamples agents while preserving their full task inventories; and two-way cluster bootstrap resamples both task and agent clusters.}
\label{tab:appendix-failure-ci-main}
\begin{tabularx}{\textwidth}{>{\raggedright\arraybackslash}p{0.23\textwidth} >{\raggedright\arraybackslash}p{0.14\textwidth} >{\raggedleft\arraybackslash}X >{\raggedleft\arraybackslash}X >{\raggedleft\arraybackslash}X >{\raggedleft\arraybackslash}X}
\toprule
Predictor & Bootstrap & AUC (95\% CI) & PR-AUC (95\% CI) & Brier (95\% CI) & ECE (95\% CI) \\
\midrule
\hcis{} & task block & 0.8481 [0.8111, 0.8799] & 0.7023 [0.6182, 0.7724] & 0.1295 [0.1180, 0.1412] & 0.1675 [0.1445, 0.1869] \\
\hcis{} & agent block & 0.8459 [0.8114, 0.8802] & 0.6990 [0.6508, 0.7528] & 0.1302 [0.1160, 0.1424] & 0.1676 [0.1412, 0.1922] \\
\hcis{} & two-way cluster & 0.8482 [0.7815, 0.8923] & 0.7042 [0.5916, 0.7947] & 0.1291 [0.1129, 0.1486] & 0.1712 [0.1374, 0.2093] \\
\hcisn{} & two-way cluster & 0.7746 [0.6994, 0.8361] & 0.5747 [0.4484, 0.6821] & 0.1868 [0.1674, 0.2052] & 0.2460 [0.1808, 0.3078] \\
Probe accuracy & two-way cluster & 0.8571 [0.8055, 0.8970] & 0.5942 [0.4847, 0.6928] & 0.1298 [0.1087, 0.1541] & 0.1541 [0.1265, 0.1837] \\
State drift rate & two-way cluster & 0.8573 [0.8121, 0.8996] & 0.5996 [0.4833, 0.6996] & 0.1299 [0.1075, 0.1543] & 0.1540 [0.1283, 0.1788] \\
\bottomrule
\end{tabularx}
\end{table*}

\begin{table*}[t]
\centering
\footnotesize
\caption{Paired two-way cluster bootstrap deltas for the main comparator pairs. Positive deltas favor \hcis{} for AUC and PR-AUC; negative deltas favor \hcis{} for Brier and ECE.}
\label{tab:appendix-paired-deltas-main}
\begin{tabularx}{\textwidth}{>{\raggedright\arraybackslash}p{0.30\textwidth} >{\raggedleft\arraybackslash}X >{\raggedleft\arraybackslash}X >{\raggedleft\arraybackslash}X >{\raggedleft\arraybackslash}X}
\toprule
Comparison & $\Delta$AUC (95\% CI) & $\Delta$PR-AUC (95\% CI) & $\Delta$Brier (95\% CI) & $\Delta$ECE (95\% CI) \\
\midrule
\hcis{} $-$ \hcisn{} & 0.0719 [0.0505, 0.0984] & 0.1260 [0.0861, 0.1732] & $-0.0561$ [$-0.0746$, $-0.0353$] & $-0.0750$ [$-0.1090$, $-0.0381$] \\
\hcis{} $-$ probe accuracy & $-0.0120$ [$-0.0258$, 0.0024] & 0.0991 [0.0704, 0.1320] & 0.0004 [$-0.0070$, 0.0075] & 0.0152 [$-0.0142$, 0.0501] \\
\hcis{} $-$ state drift & $-0.0120$ [$-0.0303$, 0.0022] & 0.0998 [0.0658, 0.1320] & 0.0004 [$-0.0077$, 0.0085] & 0.0168 [$-0.0138$, 0.0448] \\
\bottomrule
\end{tabularx}
\end{table*}

\begin{table*}[t]
\centering
\footnotesize
\caption{Trained baseline comparison. Leave-agent-out uses all other agents as training data; leave-family-out uses all other task families.}
\label{tab:appendix-trained-main}
\begin{tabularx}{\textwidth}{>{\raggedright\arraybackslash}p{0.18\textwidth} >{\raggedright\arraybackslash}p{0.18\textwidth} >{\raggedright\arraybackslash}p{0.18\textwidth} >{\raggedleft\arraybackslash}X >{\raggedleft\arraybackslash}X >{\raggedleft\arraybackslash}X >{\raggedleft\arraybackslash}X}
\toprule
Protocol & Model & Features & AUC & PR-AUC & Brier & ECE \\
\midrule
leave-agent-out & hand-specified \hcis{} & primary score & 0.8737 & 0.7625 & 0.1298 & 0.1851 \\
leave-agent-out & logistic + isotonic & audit core & 0.8694 & 0.7043 & 0.0863 & 0.0569 \\
leave-agent-out & logistic & diagnostic augmented & 0.8669 & 0.7286 & 0.0899 & 0.0758 \\
leave-family-out & hand-specified \hcis{} & primary score & 0.7648 & 0.5957 & 0.1298 & 0.1831 \\
leave-family-out & logistic & audit core & 0.8381 & 0.5454 & 0.1046 & 0.0955 \\
leave-family-out & logistic & diagnostic augmented & 0.8372 & 0.5456 & 0.1046 & 0.0955 \\
\bottomrule
\end{tabularx}
\end{table*}

\begin{table*}[t]
\centering
\footnotesize
\caption{Weight-sensitivity summary for the primary \hcis{} composition. Local perturbations multiply each weight independently before renormalization; the Dirichlet regime resamples full weight vectors around the primary confidence-free composition. The analysis treats mean absolute rank shift versus the primary ordering as the main stability statistic, while also reporting how many profiles change rank at all.}
\label{tab:appendix-weight-main}
\begin{tabularx}{\textwidth}{>{\raggedright\arraybackslash}p{0.14\textwidth} >{\raggedleft\arraybackslash}X >{\raggedleft\arraybackslash}X >{\raggedleft\arraybackslash}X >{\raggedleft\arraybackslash}X >{\raggedleft\arraybackslash}p{0.14\textwidth} >{\raggedleft\arraybackslash}p{0.12\textwidth}}
\toprule
Regime & AUC mean [5th, 95th] & PR-AUC mean [5th, 95th] & Brier mean [5th, 95th] & ECE mean [5th, 95th] & Mean abs. rank shift & Profiles with rank change \\
\midrule
Released \hcis{} & 0.8470 [0.8470, 0.8470] & 0.7000 [0.7000, 0.7000] & 0.1298 [0.1298, 0.1298] & 0.1657 [0.1657, 0.1657] & 0.0000 & 0.0 \\
Local $\pm 10\%$ & 0.8472 [0.8465, 0.8483] & 0.7001 [0.6979, 0.7020] & 0.1299 [0.1288, 0.1311] & 0.1666 [0.1613, 0.1735] & 0.3060 & 8.4 \\
Local $\pm 25\%$ & 0.8477 [0.8443, 0.8523] & 0.7009 [0.6950, 0.7089] & 0.1302 [0.1277, 0.1334] & 0.1687 [0.1566, 0.1842] & 0.8659 & 15.0 \\
Dirichlet & 0.8485 [0.8438, 0.8547] & 0.7011 [0.6933, 0.7106] & 0.1305 [0.1263, 0.1368] & 0.1691 [0.1508, 0.1906] & 1.1187 & 16.7 \\
\bottomrule
\end{tabularx}
\end{table*}

\begin{table*}[t]
\centering
\footnotesize
\caption{Feature-availability and leakage audit for the mainline and exploratory integrity compositions.}
\label{tab:appendix-feature-main}
\begin{tabularx}{\textwidth}{>{\raggedright\arraybackslash}p{0.19\textwidth} >{\raggedright\arraybackslash}p{0.18\textwidth} >{\centering\arraybackslash}p{0.10\textwidth} >{\centering\arraybackslash}p{0.10\textwidth} >{\raggedright\arraybackslash}X}
\toprule
Signal & Availability stage & Uses gold probe label? & Uses final outcome? & Interpretation in the benchmark \\
\midrule
State drift & post-probe & Yes & No & Core diagnostic signal after answer matching; included in \hcis{}, \hcisfull{}, and \hcisn{} but not a pre-probe feature. \\
Response-confidence risk & post-probe & No & No & Agent-reported confidence transformed into risk; excluded from \hcis{} and retained in \hcisfull{} and \hcisn{}. \\
Uncertainty risk & post-probe & No & No & Binary entropy of response confidence; excluded from \hcis{} because cross-provider confidence comparability is limited. \\
Trajectory risk & post-probe & No & No & Derived from probe-side trajectory diagnostics and included in \hcis{}, \hcisfull{}, and \hcisn{}. \\
Context-position risk & pre-probe metadata & No & No & Derived only from probe position, phase, and distractor status; included in \hcis{}, \hcisfull{}, and \hcisn{}. \\
Success-probability risk & post-probe & No & No & Outcome-adjacent self-assessment proxy; excluded from \hcis{} and retained only in \hcisn{}. \\
Task success oracle & post-outcome & No & Yes & Included only as an upper-bound diagnostic, never as an early-warning signal derived from the benchmark itself. \\
\bottomrule
\end{tabularx}
\end{table*}

\begin{table*}[t]
\centering
\scriptsize
\caption{Cross-provider confidence-comparability audit for the primary \hcis{} and legacy confidence-sensitive variants. The audit treats confidence and uncertainty terms as legacy heuristic components rather than universally comparable measurements across local and gateway profiles.}
\label{tab:appendix-confidence-main}
\setlength{\tabcolsep}{2pt}
\begin{tabular}{@{}>{\raggedright\arraybackslash}p{0.30\textwidth} >{\raggedleft\arraybackslash}p{0.075\textwidth} >{\raggedleft\arraybackslash}p{0.08\textwidth} >{\raggedleft\arraybackslash}p{0.075\textwidth} >{\raggedleft\arraybackslash}p{0.075\textwidth} >{\raggedleft\arraybackslash}p{0.095\textwidth} >{\raggedleft\arraybackslash}p{0.095\textwidth}@{}}
\toprule
Variant & AUC & PR-AUC & Brier & ECE & Rank shift & Rank changes \\
\midrule
Primary \hcis{} & 0.8469 & 0.6992 & 0.1298 & 0.1657 & 0.0000 & 0 \\
Legacy \hcisfull{} & 0.7996 & 0.6409 & 0.1642 & 0.2223 & 9.9375 & 32 \\
Probe-accuracy proxy & 0.8587 & 0.5983 & 0.1297 & 0.1528 & 4.1875 & 26 \\
Drift-only comparator & 0.8587 & 0.5983 & 0.1297 & 0.1528 & 4.1875 & 26 \\
Legacy \hcisfull{} with per-agent normalized confidence & 0.8381 & 0.6785 & 0.1427 & 0.2037 & 5.5625 & 29 \\
\bottomrule
\end{tabular}
\end{table*}

\begin{figure*}[t]
\centering
\includegraphics[width=\textwidth]{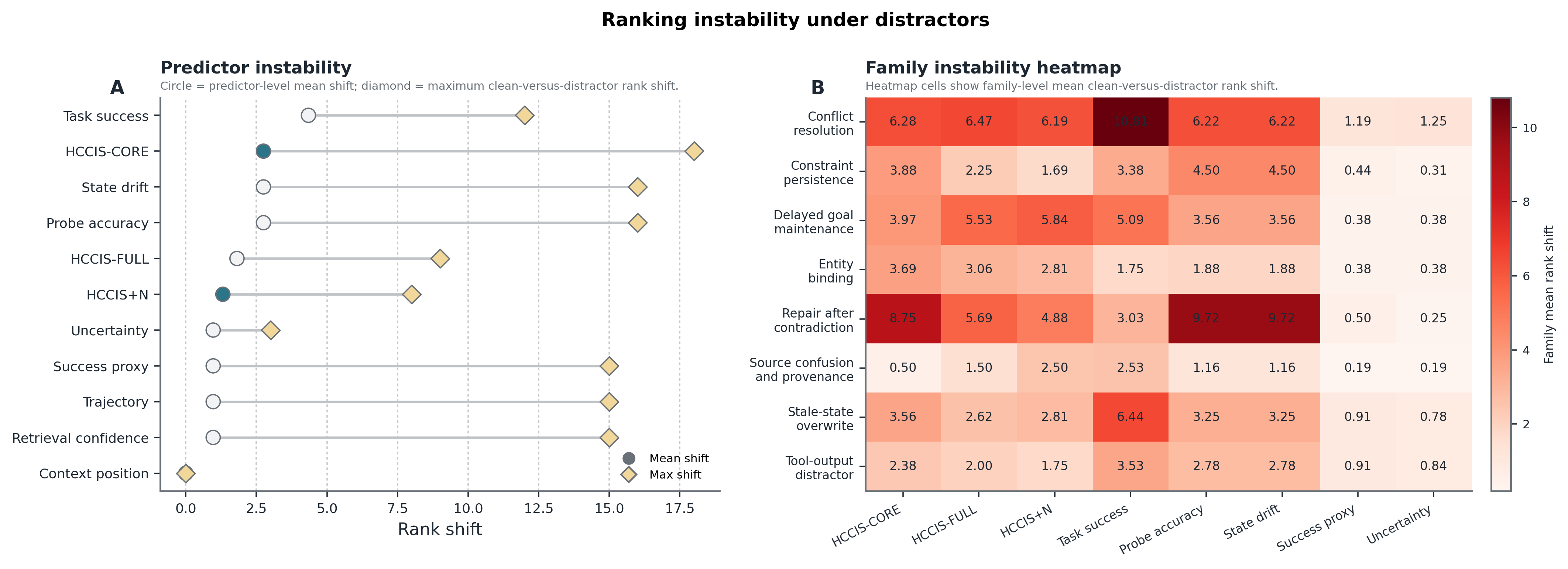}
\caption{Ranking instability under distractors for the expanded 32-profile grid. Panel A compares predictor-level mean and maximum clean-to-distractor rank movement, and Panel B reports family-level mean rank shift, showing that success rankings move much more than integrity rankings in commitment-heavy families.}
\label{fig:appendix-instability-main}
\end{figure*}

\begin{figure*}[t]
\centering
\includegraphics[width=\textwidth]{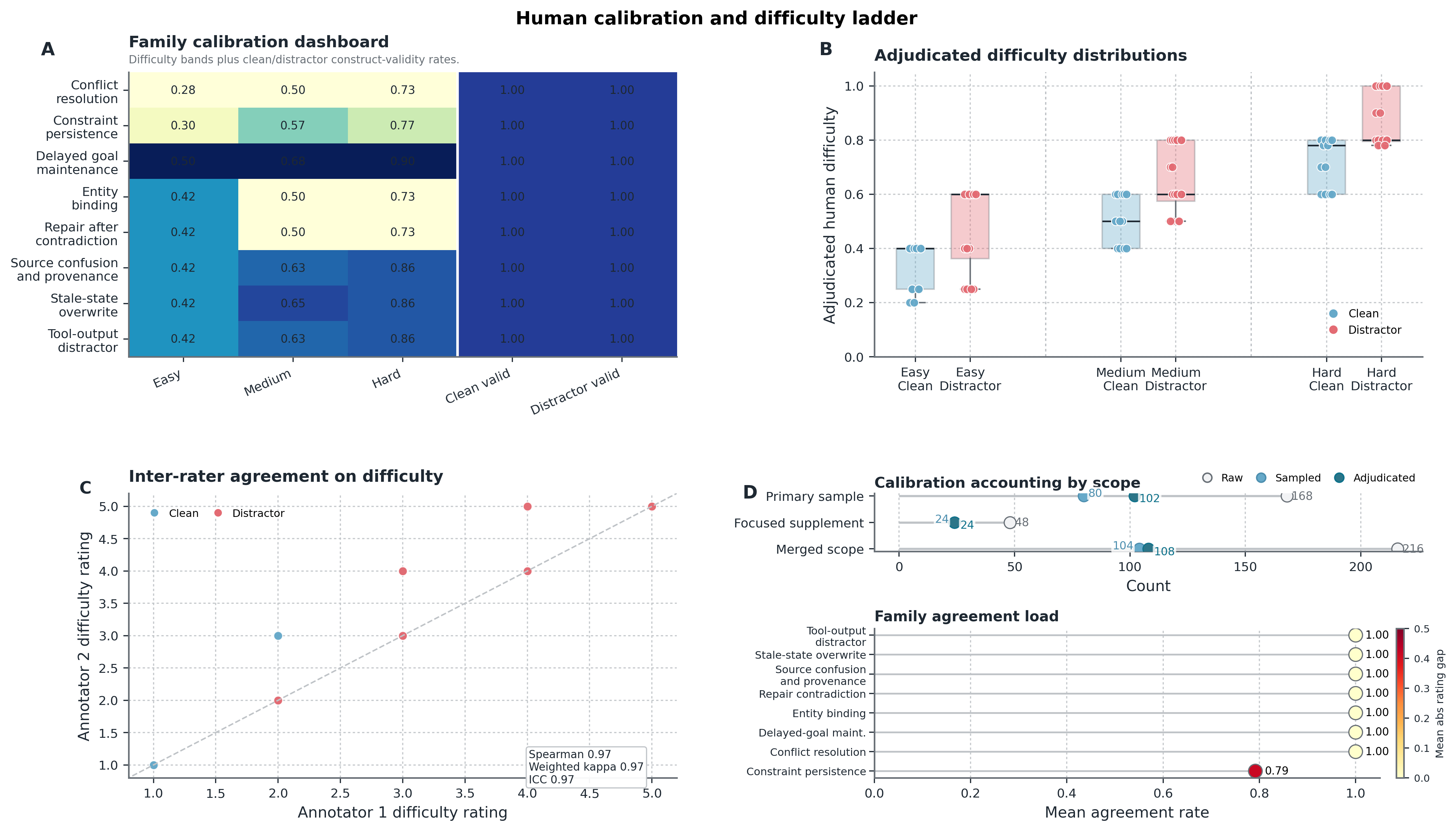}
\caption{Human calibration and construct-validity dashboard. The appendix preserves the full calibration visualization, including inter-rater agreement and reporting-scope accounting, while the primary analysis uses only the final merged counts and monotonicity facts needed for the benchmark-level conclusions.}
\label{fig:appendix-calibration-main}
\end{figure*}

\begin{figure*}[t]
\centering
\includegraphics[width=\textwidth]{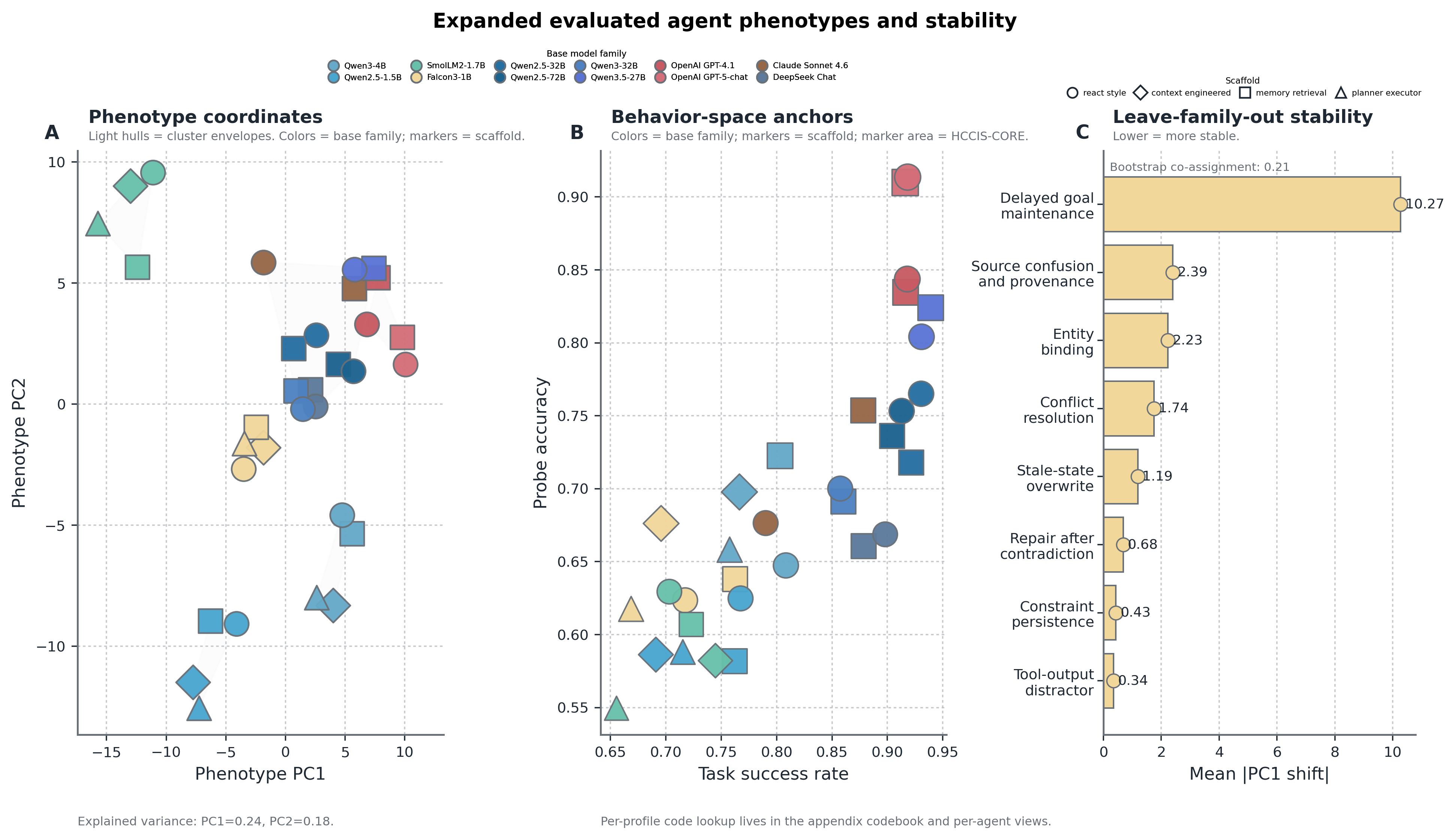}
\caption{Supporting phenotype map and leave-family-out stability for the expanded 32-profile evaluated grid. Panel A shows phenotype coordinates with light cluster envelopes; Panel B shows behavior-space anchors with colors for base model families, markers for scaffolds, and marker area proportional to \hcis{}; Panel C reports leave-family-out stability. Compact profile-code lookup remains in the appendix codebook and per-agent views.}
\label{fig:appendix-phenotype-main}
\end{figure*}

\begin{figure*}[t]
\centering
\includegraphics[width=\textwidth]{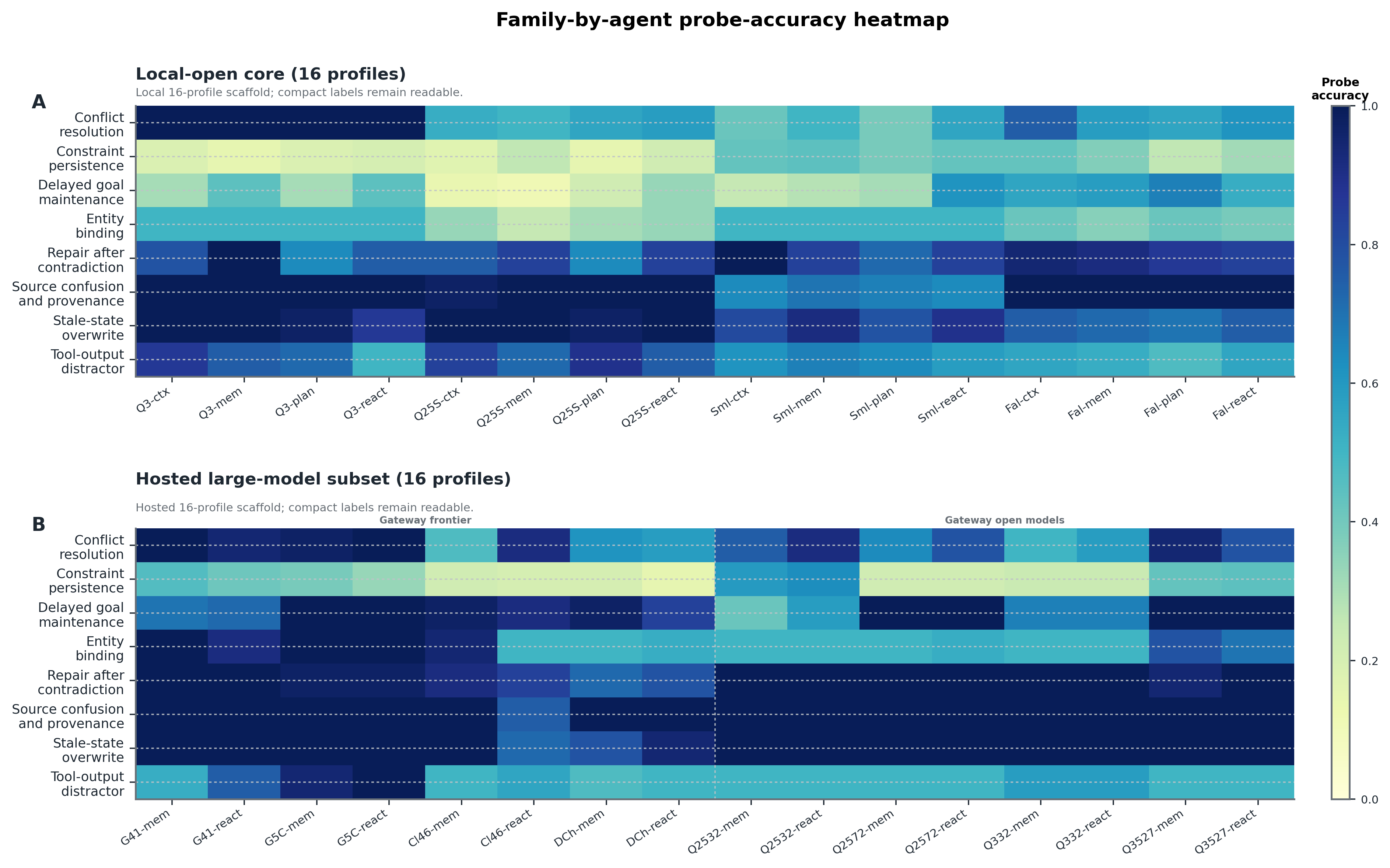}
\caption{Family-by-agent probe-accuracy heatmap for the expanded evaluated grid. The full 32-profile view is split into the local 16-profile subset in the top panel and the hosted 16-profile subset, with 8 gateway-frontier and 8 gateway-open profiles, in the bottom panel.}
\label{fig:appendix-agent-breakdown-main}
\end{figure*}

\begin{figure*}[t]
\centering
\includegraphics[width=\textwidth]{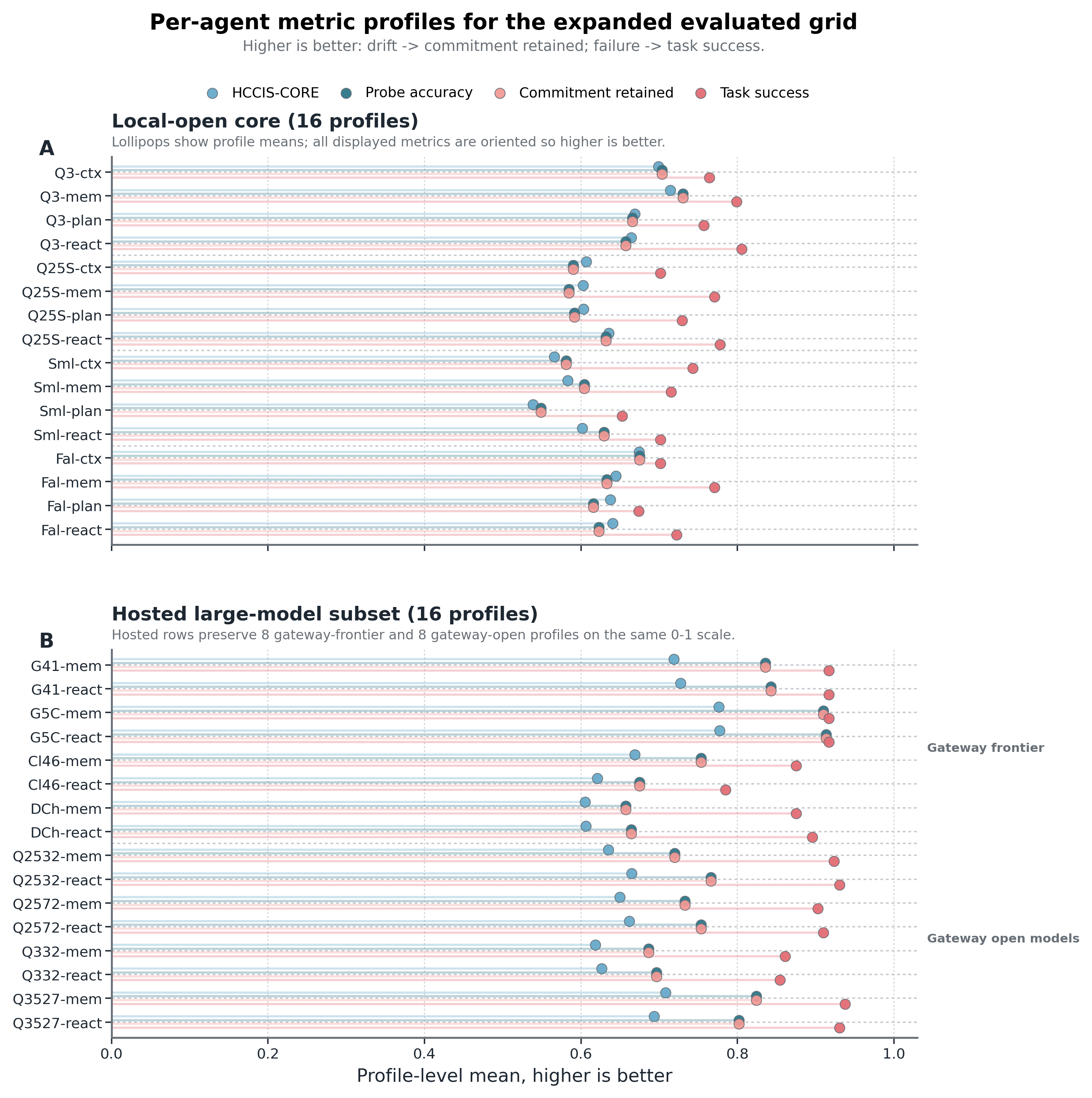}
\caption{Per-agent metric profiles for the expanded evaluated grid. The split appendix layout keeps all 32 profiles visible as local and hosted panels. Values are plotted as profile-level lollipops on a common higher-is-better scale: \hcis{} and probe accuracy are shown directly, while state drift is shown as commitment retained ($1-\mathrm{drift}$) and task failure as task success ($1-\mathrm{failure}$).}
\label{fig:appendix-agent-profiles-main}
\end{figure*}

\begin{figure*}[t]
\centering
\includegraphics[width=\textwidth]{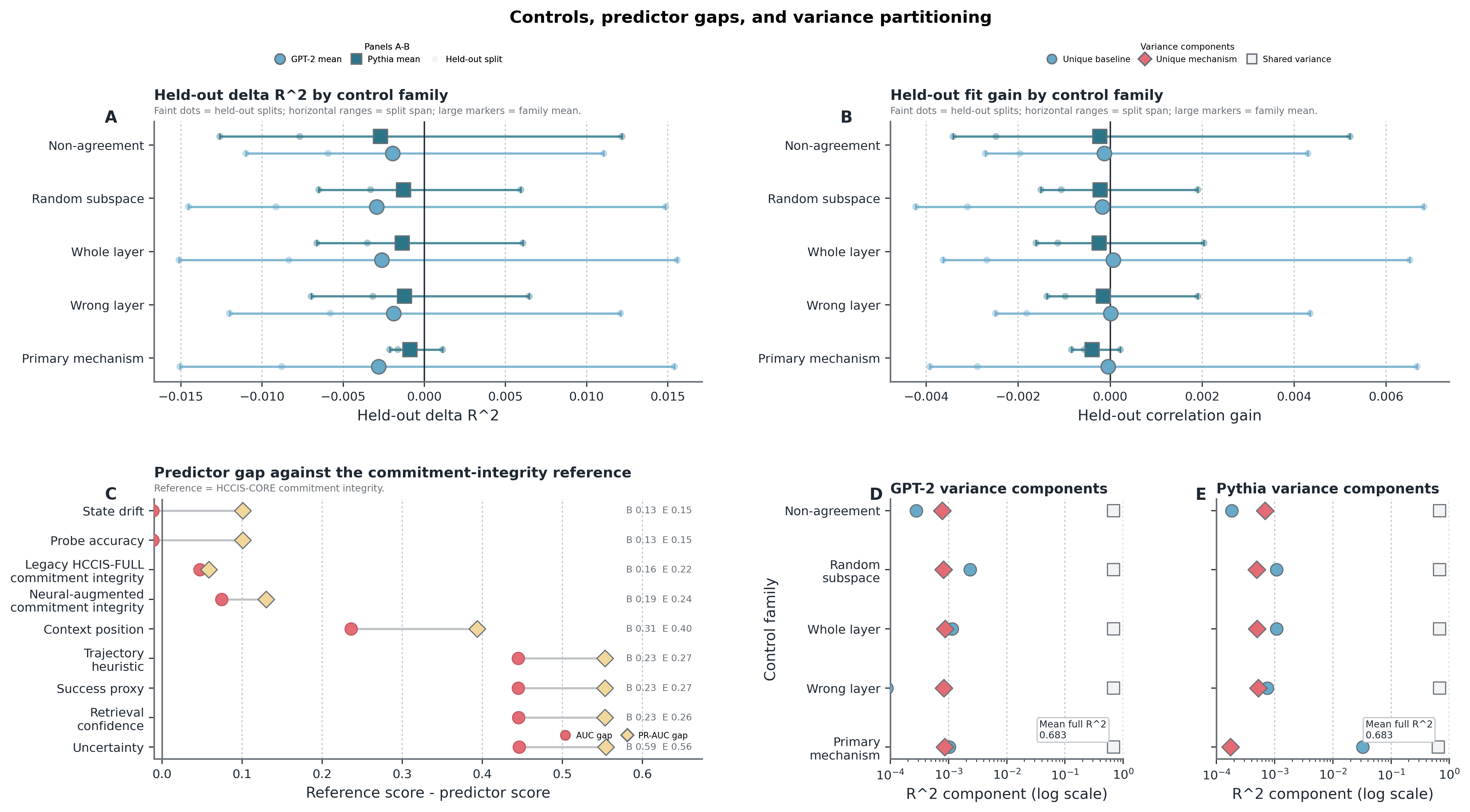}
\caption{Supporting diagnostics for controls, predictor gaps, and variance partitioning. Panels A--B layer faint held-out split dots, large family means, and horizontal held-out split ranges for each control family; Panel C shows AUC and PR-AUC gaps against the commitment-integrity reference; and Panels D--E show GPT-2 and Pythia variance components. The negative-control view shows numerically small unique mechanistic contributions and dominant shared variance in the evaluated fits.}
\label{fig:appendix-controls-main}
\end{figure*}

\begin{figure*}[t]
\centering
\includegraphics[width=\textwidth]{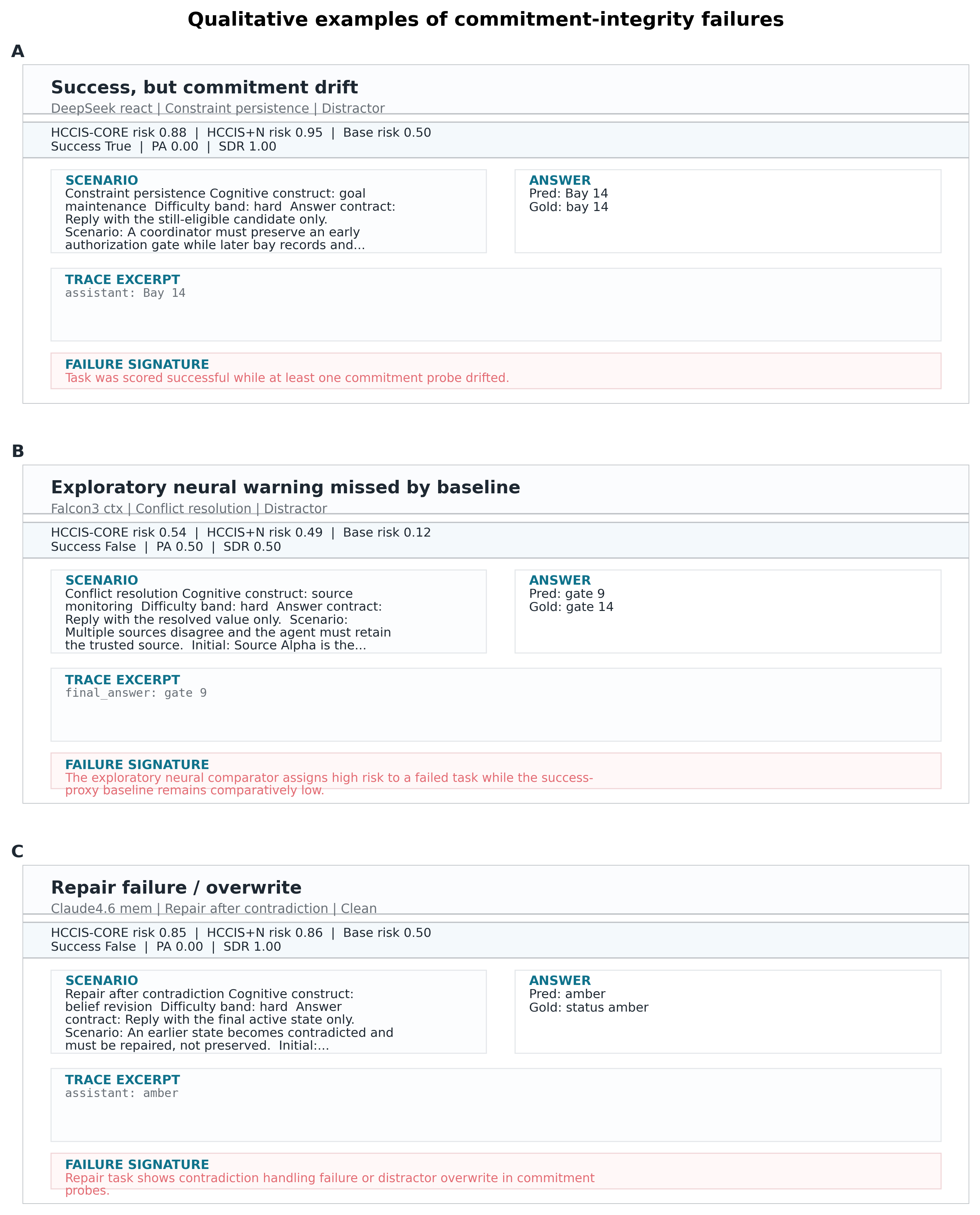}
\caption{Supplementary qualitative case cards automatically selected from the case-study pipeline. The cards are stacked vertically to keep scenario, answer, trace excerpt, and failure-signature text readable, and the examples are generated from saved benchmark outputs rather than manually authored after inspection.}
\label{fig:supp-cases}
\end{figure*}

\begin{figure*}[t]
\centering
\includegraphics[width=\textwidth]{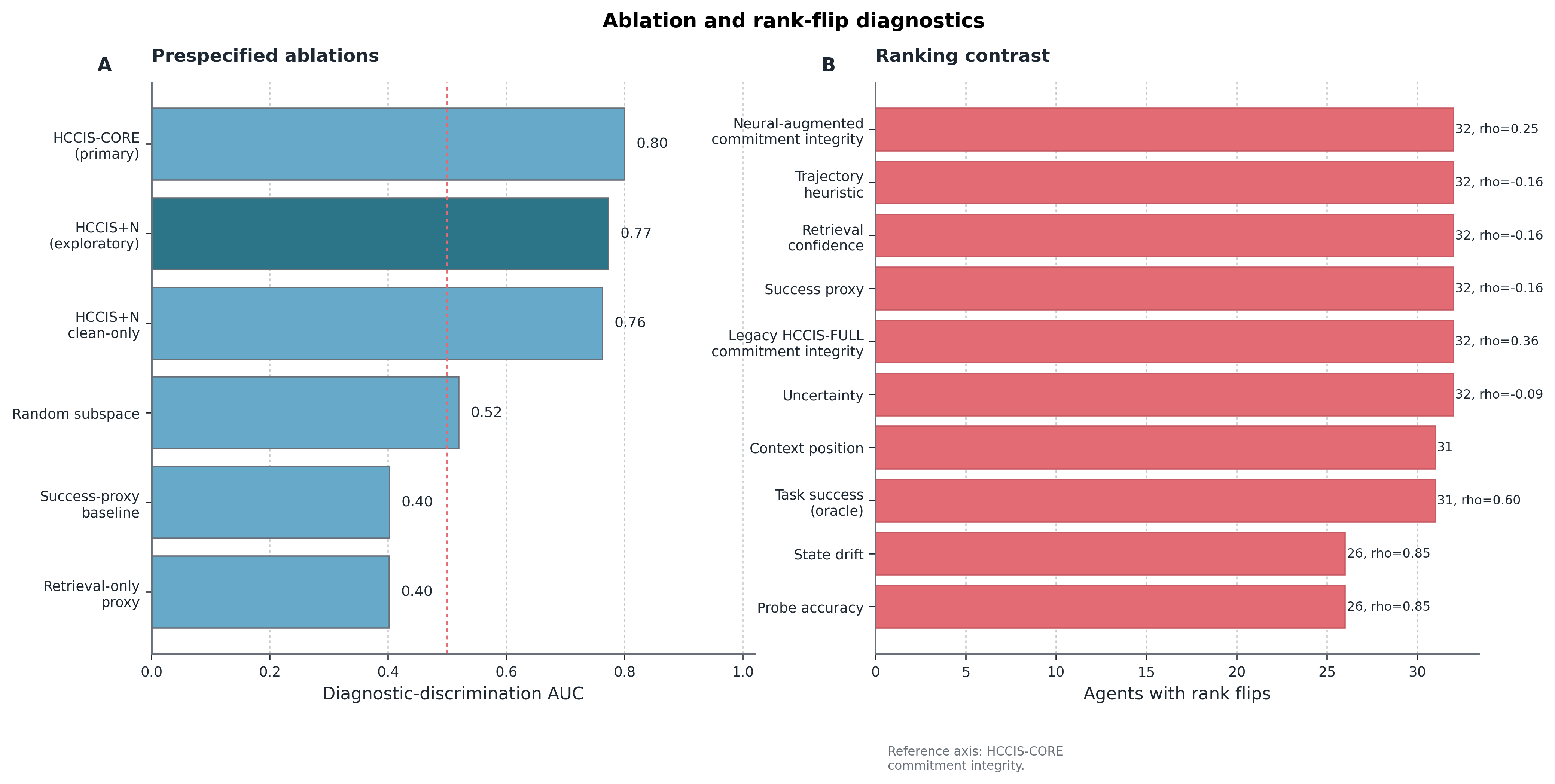}
\caption{Supplementary ablation and ranking-flip diagnostics. Panel A summarizes the preregistered ablation-style controls associated with the benchmark artifacts. Panel B compares ranking-flip counts and correlations against the reference integrity axis.}
\label{fig:supp-ablation}
\end{figure*}

These appendix figures connect aggregate summaries to concrete benchmark records. They show item-level failures and ablation behavior that support the benchmark-level conclusions through inspectable instances and controls.

\bibliographystyle{plainnat}
\bibliography{references}

\end{document}